\pdfoutput=1

\documentclass[11pt]{article}

\usepackage[final]{acl}
\usepackage{amsmath}
\usepackage{amssymb}
\usepackage{bbm}
\usepackage{booktabs}
\usepackage{times}
\usepackage{latexsym}
\usepackage[normalem]{ulem}
\usepackage{pifont}

\usepackage{multirow}

\usepackage[T1]{fontenc}

\usepackage[utf8]{inputenc}

\usepackage{microtype}

\usepackage{inconsolata}

\usepackage{graphicx}

\title{FineCops-Ref: A new Dataset and Task for Fine-Grained Compositional Referring Expression Comprehension}

\author{
 \textbf{Junzhuo Liu\textsuperscript{1}},
 \textbf{Xuzheng Yang\textsuperscript{1}},
 \textbf{Weiwei Li\textsuperscript{1}},
 \textbf{Peng Wang\textsuperscript{1\thanks{Corresponding author.}}}
 \\
 \textsuperscript{1}University of Electronic Science and Technology of China
 \\
 \texttt{junzhuo.cs@gmail.com}, \texttt{yangxuzheng@std.uestc.edu.cn},\\
 \texttt{davelee.uestc@gmail.com}, \texttt{wangpeng8619@gmail.com}
 }

\begin{document}
\maketitle
\begin{abstract}
Referring Expression Comprehension (REC) is a crucial cross-modal task that objectively evaluates the capabilities of language understanding, image comprehension, and language-to-image grounding. Consequently, it serves as an ideal testing ground for Multi-modal Large Language Models (MLLMs). In pursuit of this goal, we have established a new REC dataset characterized by two key features: Firstly, it is designed with controllable varying levels of difficulty, necessitating multi-level fine-grained reasoning across object categories, attributes, and multi-hop relationships. Secondly, it includes negative text and images created through fine-grained editing and generation based on existing data, thereby testing the model's ability to correctly reject scenarios where the target object is not visible in the image—an essential aspect often overlooked in existing datasets and approaches. Utilizing this high-quality dataset, we conducted comprehensive evaluations of both state-of-the-art specialist models and MLLMs. Our findings indicate that there remains a significant gap in achieving satisfactory grounding performance. We anticipate that our dataset will inspire new approaches to enhance visual reasoning and develop more advanced cross-modal interaction strategies, ultimately unlocking the full potential of MLLMs. Our code and the datasets are available at \url{https://github.com/liujunzhuo/FineCops-Ref}.
\end{abstract}

\section{Introduction}
Despite significant advancements in multimodal large language models (MLLMs), a critical challenge remains in ensuring these models' responses are grounded in visual content
rather than solely derived from linguistic cues~\cite{tong2024eyes,zhai2023halle-switch,miyai2024unsolvable}. Vision-language models (VLMs) often treat language as a bag of words, lacking meaningful engagement with word order, attributes, or relationships~\cite{ma2023crepe,thrush2022winoground,tong2024eyes,yuksekgonul2022and}, and exhibit poor grounding and spatial reasoning abilities~\cite{chen2024spatialvlm,tong2024eyes,zhang-etal-2024-countercurate}.

Current evaluation methods utilize Visual Question Answering or Image-Text Retrieval to evaluate the compositional reasoning or
grounding abilities of MLLMs. However, these methods provide an indirect assessment of the models' visual grounding capabilities. In contrast, the Referring Expression Comprehension (REC) task requires a model to directly output the bounding box coordinates of a target object based on a given language expression, serving as an ideal testing ground for MLLMs.

Recent MLLMs, leveraging substantial grounding data~\cite{chen2023shikra,wang_visionllm_2023, wang2023cogvlm} and specifically designed visual modules~\cite{you2024ferret,li2024covlm}, have achieved impressive results on common REC benchmarks like RefCOCO/+/g~\cite{yu2016modeling}. However, these benchmarks lack considerations of compositional reasoning, allowing models to perform well without understanding linguistic structure or even without the expression~\cite{cirik-etal-2018-visual,akula-etal-2020-words}. Additionally, current VLMs struggle with negative samples, where the target object is absent from the image~\cite{chen2020cops,kurita2023refego,you2024ferret}. This limitation is further exacerbated by the lack of robustness in existing datasets, which fail to provide the necessary complexity and variability to thoroughly evaluate MLLMs.

\textbf{In response, we introduce FineCops-Ref}, a benchmark specifically designed to address these limitations. Our dataset introduces controlled difficulty levels, compelling MLLMs to perform fine-grained reasoning across object categories, attributes, and multi-hop relationships. We classify the difficulty levels based on the number of attributes and relationships necessary for locating the target object. For instance, if there is only one possible target in the image, the difficulty level is 1 regardless the complexity of the expression. If the model needs to understand at least two or more relationships and attribute information, the difficulty level is 3. Moreover, FineCops-Ref incorporates negative samples crafted through meticulous editing, testing the models' resilience against misalignments and hallucinations, thereby assessing their true visual grounding capabilities.

Our comprehensive evaluation with state-of-the-art models reveals a significant gap in grounding performance, highlighting the need for advanced visual reasoning strategies. We present several core findings in our study. Firstly, for simple REC tasks with a difficulty level 1, traditional vision-language models, despite their relatively smaller parameter sizes, maintained a significant advantage. Secondly, all models exhibited poorer performance at difficulty levels greater than 1, while MLLMs demonstrated stronger capabilities under these conditions. In terms of negative data, all models showed weak performance, even in the simplest scenarios where the image does not contain an object matching the category specified in the expression. Additionally, we observed a positive correlation between precision on positive samples and recall with negative samples, with traditional vision-language models and MLLMs displaying different tendencies.

To enhance the fine-grained compositional reasoning capabilities of existing models, we employed the same pipeline used to construct our benchmark to create a rich training dataset that includes both positive and negative samples. Fine-tuning on this training dataset significantly improved model performance, with further improvements observed on the RefCOCO/+/g dataset. We make FineCops-Ref and the code for our data generation pipeline publicly available under the CC BY 4.0 License.

\section{Related Works}

\textbf{Referring expression comprehension.}
The REC methods can generally be divided into two categories based on whether or not it uses LLMs: specialist and MLLMs. Specialists typically extract text and image features separately and perform multi-stage fusion~\cite{liu2023grounding,yan2023universal,kamath2021mdetr}. Their training tasks often include various object location tasks. Recently, \citet{Zhao_2024_CVPR} achieved excellent results on two visual grounding (VG) benchmarks by leveraging hard negative samples in training. 

On the other hand, MLLMs directly input the projected visual features into the LLM.
Recent methods aim to enhance grounding capabilities in MLLMs through dataset construction with coordinate information and additional visual modules. Common methods for datasets include transforming traditional visual datasets into an instruction-following format using templates~\cite{li-etal-2024-groundinggpt,pramanick2023jack,wang_visionllm_2023}, correlating object coordinates with existing captions~\cite{peng2024grounding,qi2024cogcom}, and using LLMs to generate grounded question-answer pairs based on images, object coordinates, and captions~\cite{you2024ferret, wang2024the}. 

In terms of visual modules, some methods integrate additional visual components, such as GLaMM~\cite{Rasheed_2024_CVPR} and LLaVA-Grounding~\cite{zhang2023llava-g}, while others extract regional features as additional inputs~\cite{ma2024groma,shao2024visualcot,you2024ferret,li2024covlm}.

\begin{figure*}[t]
  \centering
  \includegraphics[width=1\textwidth]{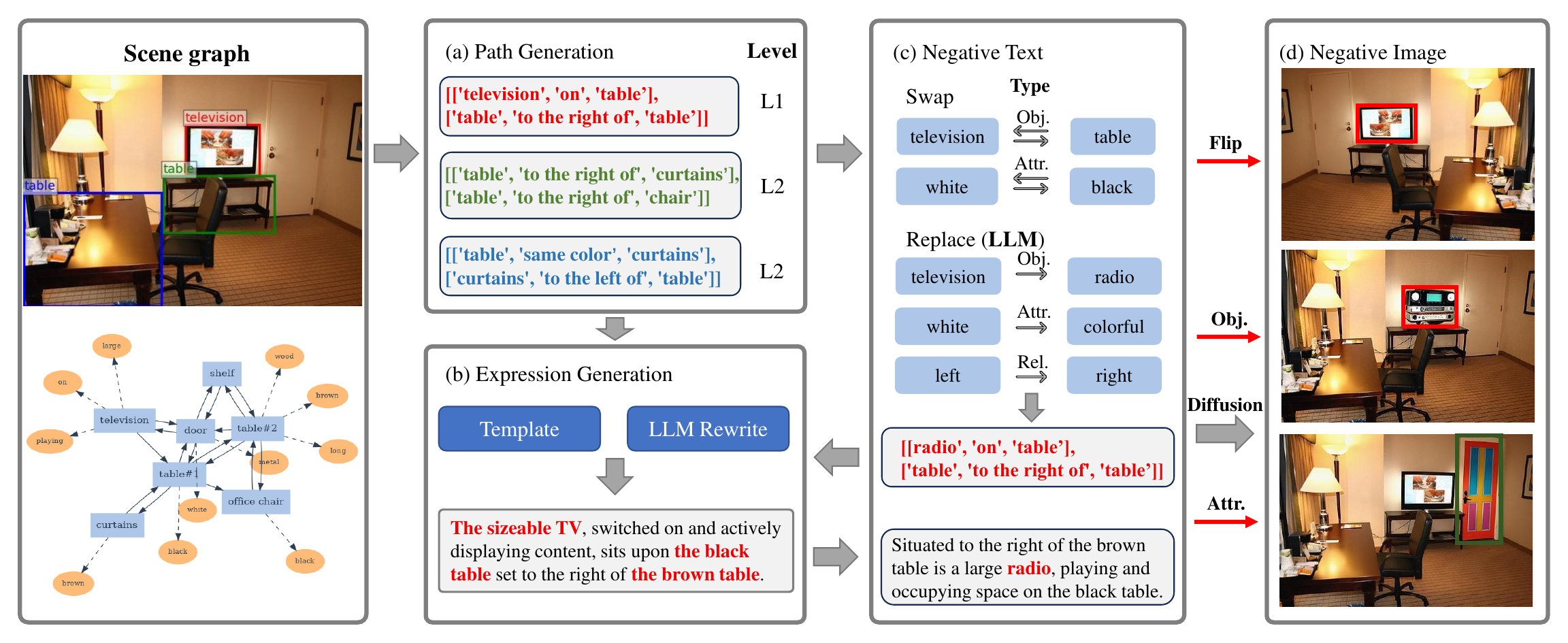}
  \caption{The data construction pipeline of FineCops-Ref. Given an image, we first generate paths based on its scene graph. Then, we fill paths into templates and obtain the positive referring expression through LLM rewriting. Meanwhile, we utilize LLM to generate negative expressions, and based on this, we employ diffusion model to create fine-grained editing negative images.}
  \label{pipeline}
\end{figure*}

\textbf{Evaluation of Compositional Reasoning.}
Current multimodal models, including advanced MLLMs like GPT-4V, exhibit poor compositional reasoning, often treating language as a bag of words without considering word order, attributes, or relationships between objects~\cite{suhr-etal-2019-corpus,ma2023crepe,diwan-etal-2022-winoground,tong2024eyes,yuksekgonul2022and}. Common evaluation benchmarks involve constructing hard negative captions to test models' capabilities, such as distinguishing between "a mug in some grass" and "some grass in a mug"~\cite{parcalabescu-etal-2022-valse,thrush2022winoground,ma2023crepe}. \citet{hsieh_sugarcrepe_2023} found that previous benchmarks have language biases and that a simple grammar model can distinguish negative captions. Some benchmarks focus on negative images~\cite{ray_cola_2023,yarom_what_2023,zhang-etal-2024-countercurate,NEURIPS2023_coco_counter}, while others primarily focus on spatial relationships~\cite{zhang-etal-2024-countercurate,liu-etal-2023-visual,yang2019spatialsense,chen2024spatialvlm}.

For REC tasks, \citet{akula-etal-2020-words} critically examined RefCOCOg, showing that 83.7\% of test instances do not require reasoning on linguistic structure, and proposed the Ref-Adv dataset, which perturbs original expressions to refer to different target objects. CLEVR-Ref+~\cite{liu2019clevr} is a synthetic dataset emphasizing relationships, attributes, and linguistic logic. 
Cops-Ref~\cite{chen2020cops} and Ref-Reasoning~\cite{yang2020graph} use GQA scene graphs~\cite{hudson2019gqa} and rule-based methods to create large-scale compositional referring expression comprehension datasets in real-world scenarios. Cops-Ref additionally added distracting images based on attributes, relationships, and target names.  GITM-MR~\cite{Wu_2023_ICCV_GITM-MR} explores mismatched relationship in the REC task. RefEgo~\cite{kurita2023refego} and OmniLabel~\cite{schulter2023omnilabel} consider out-of-distribution scenarios where referred targets do not exist in the image.

This paper addresses the limitations of previous benchmarks by constructing a REC dataset that comprehensively evaluates the compositional understanding abilities of existing multimodal models.

\section{FineCops-Ref}
FineCops-Ref includes both positive and negative data. Figure~\ref{pipeline} illustrates the dataset construction pipeline.

\subsection{Creating Positive Data}

\textbf{Path Generation.}
We employ image scene graphs from GQA~\cite{hudson2019gqa} for path generation. The scene graphs contain detailed information about objects, attributes, and relations. To ensure accuracy, we first filter the objects based on their suitability as target or related objects. We leverage annotations from InstInpaint~\cite{yildirim2023inst} and apply additional filters such as keywords and object size. Next, we generate several paths for each of the filtered objects, as show in Figure~\ref{pipeline}(a).
To eliminate ambiguity, we utilize unique attributes or relations to identify the target object that shares the same category as other objects in the image to ensure that every generated path is unique.

\textbf{Data categorization.}
We categorize the paths into three difficulty levels based on the complexity of fine-grained reasoning. Level 1 indicates that there are no other objects in the image belonging to the same category as the target object, such as the TV in Figure~\ref{pipeline}. In this case, the model can locate the target without requiring contextual understanding. Level 2 signifies the presence of another object with the same category as the target in the image, where the target can be distinguished through one unique attribute or relation. Level 3 requires at least two or more relationships and attribute information. The difficulty levels are determined by the intricacy of fine-grained reasoning, rather than the complexity of the textual description. 

\begin{table*}[t]
\centering
\resizebox{\textwidth}{!}{
\begin{tabular}{lcccccccc} 
\toprule
                   &                   &                           &                           &                   &                    & \textbf{Positive}   & \multicolumn{2}{c}{\textbf{Negative}}  \\ 
\cmidrule(lr){7-7}
\cmidrule(lr){8-9}
\textbf{Benchmark} & \textbf{Unconstrained} & \textbf{Cops.} & \textbf{Difficulty level} & \textbf{Neg. text} & \textbf{Neg. image} & \textbf{Expression} & \textbf{Expression} & \textbf{Image}    \\ 
\midrule
RefCOCO            & \ding{51}                 &                           &                           &                   &                    & 10,752               & -                   & -                 \\
RefCOCO+           & \ding{51}                 &                           &                           &                   &                    & 10,615               & -                   & -                 \\
RefCOCOg           & \ding{51}                 &                           &                           &                   &                    & 9,602               & -                   & -                 \\
Ref-reasoning      &                   & \ding{51}                         & \ding{51}                         &                   &                    & 34,609              & -                   & -                 \\
Cops-ref           &                   & \ding{51}                         &                           &                   & \ding{51}                  & 12,586               & -                   & 37,758             \\
Ref-adv            & \ding{51}                 & \ding{51}                         & \ding{51}                         & \ding{51}                 &                    & 9,602                & 3,704                & -                 \\
Ours                & \ding{51}                 & \ding{51}                         & \ding{51}                         & \ding{51}                 & \ding{51}                  & 9,605                & 9,814                & 8,507              \\
\bottomrule
\end{tabular}}
\caption{\label{benchmark_comparison}
Comparison between the proposed benchmark and other REC benchmarks. Unconstrained  indicates the final expression is not constrained by the templates. Cops. indicates compositional reasoning. On the right hand side, the test set count of each benchmark is listed.
}
\end{table*}

\textbf{Expression Generation.}
To mitigate potential biases in the scene graphs, we first apply frequency-based sampling of relationships, attributes, and object categories along the generated paths. Subsequently, we use predefined templates to generate referring expressions. The details of predefined templates are provided in Appendix~\ref{sec:appendix_dataset_0}.

To further augment the naturalness and diversity of these expressions, we leverage GPT-3.5-turbo to rewrite the referring expressions. By incorporating well-designed instructions and examples, we achieve a wider range of linguistically diverse and natural expressions. The prompts used for rewriting are listed in Appendix~\ref{sec:appendix_dataset_3}.

\textbf{Human Filter.}
Due to inherent limitations in the scene graph annotations, the generated paths, particularly for Level 2 and Level 3, may contain inaccuracies, leading to non-unique target references. To address this, human annotators manually filtered the test set. Further details can be found in Appendix~\ref{sec:appendix_dataset_4}.

\subsection{Generating Negative Data}
To conduct a thorough and systematic assessment of REC of existing VLMs, we generate hard negatives from both textual and visual sources. 
Similar to positive data, negative data are categorized into different levels based on difficulty. Negative Level 1 involves alterations to the target object in the expressions, which are relatively straightforward for the model to identify. Level 2 involves modifications to the related objects, disrupting the contextual information and posing a greater challenge for existing models to recognize.

\textbf{Generating Negative Expressions.}
Our set of negative expressions encompasses a wide range of challenging types. Inspired by CREPE~\cite{ma2023crepe} and SUGARCREPE~\cite{hsieh_sugarcrepe_2023}, we consider various forms of hard negatives. In total, FineCops-Ref covers 5 fine-grained types of hard negative expressions. These types can be broadly classified into two categories: Replace and Swap. The Replace category involves generating negative expressions by substituting a portion of the original expression, whether it is an object, attribute, or relation. We utilize LLM to determine the most appropriate negative word, ensuring that the negative expression is genuinely negative while only slightly deviating from the original expression. We experimented with various replacement methods and found that LLM-based replacements performed the best, as discussed in Appendix~\ref{sec:appendix_dataset_2}. The Swap category entails generating negative expressions by interchanging two attributes or objects within the same category. We further employ LLM to rewrite these new expressions.

\textbf{Generating Negative Images.} We consider the necessity of negative images from the following aspects. First, negative images enables a more thorough assessment of models' visual parsing capabilities. Additionally, evaluations conducted by Visualgptscore~\cite{lin2023visualgptscore} suggest that negative expressions may lack plausibility and fluency and can be detected by language prior.

We generate hard negative images with subtle differences from the original, such as modifications to objects, attributes, or relations. For simple positional relationships, we employ horizontal flips. For more intricate modifications involving objects and attributes, we utilize PowerPaint~\cite{zhuang2023task}, an exceptional image inpainting model offering versatility, to perform precise edits on the images. To guide PowerPaint in editing the image, we utilize LLM-generated replacements as textual guides and the bounding boxes as masks. Overall, FineCops-Ref encompasses 5 distinct types of challenging negative images. Further details on the data types, along with example expressions and images, can be found in Appendix~\ref{sec:appendix_dataset_1}.

\textbf{Negative Data Debiasing.}
During the generation of negative samples, some implausible or incoherent expressions, as well as unreasonable and easily distinguishable negative images, are inevitable. We employed several techniques to filter out unsuitable samples and improve the quality of the benchmark. For negative expressions, we employ the Adversarial Refinement technique proposed by SUGARCREPE~\cite{hsieh_sugarcrepe_2023}, which helps mitigate biases and unintended artifacts in the dataset.

To exclude inappropriate and excessively unreasonable negative images, we employ a multi-step filtering process. First, we use CLIP~\cite{radford2021learning} to ensure that the similarity between the negative text and the positive image is lower than the similarity between the positive text and the positive image. Next, we apply the diffusion-generated inspection model DIRE~\cite{wang2023dire} to filter out excessively unnatural images, excluding those with scores above 0.2. Finally, we use DINOv2~\cite{oquab2023dinov2} to compute the image-image similarity between the positive image and the generated negative images, retaining the candidate negative image with the highest DINOv2 score from a set of 10 candidates, thereby minimizing noise.

\begin{table}
\centering
\resizebox{\columnwidth}{!}{
\begin{tabular}{cccc} 
\toprule
\textbf{Set}   & \textbf{Positive} & \textbf{Negative expression} & \textbf{Negative image}  \\ 
\midrule
Train & 163,792   & 80,451               & -               \\
Val   & 18,455    & 9,029                & -               \\
Test  & 9,605     & 9,814                & 8,507            \\
\bottomrule
\end{tabular}}
\caption{\label{dataset_statics_all}
Dataset statistics.}
\end{table}

\subsection{Dataset Statistics}
FineCops-Ref consists of 9,605 positive expressions, 9,814 negative expressions, and 8,507 negative images in test set. Table~\ref{benchmark_comparison} provides a comparison between FineCops-Ref and other visual grounding benchmarks. FineCops-Ref combines the advantages of unconstrained expression, fine-grained compositional reasoning, difficulty level, and hard negatives at both textual and visual levels. Additionally, we partition the training set and validation set simultaneously as in Table~\ref{dataset_statics_all}. For more statistics, please refer to the Appendix~\ref{sec:appendix_dataset_1}.

\subsection{Metrics}

To evaluate performance on positive data, we use the common metric Precision@k.  
When both positive and negative data are present in the test set, we treat the negative samples as distractors for the positive samples, and introduce two additional metrics:

\textbf{Recall@k}:  
We treat the REC task as a bounding box retrieval problem. For each negative sample paired with its corresponding positive sample, we first obtain the predicted bounding boxes from the model, along with their confidence scores for both positive and negative samples. These bounding boxes are then ranked based on their confidence scores.  
Recall@k measures the proportion of negative-positive pairs where at least one of the top \(k\) predicted bounding boxes has an IoU greater than 0.5 with the ground truth bounding box. It specifically assesses the model's ability to avoid assigning high confidence scores to negative samples. Formally, Recall@k is defined as:

{\small
\begin{equation}
\text{Recall@k} = \frac{1}{N} \sum_{i=1}^{N} \mathbbm{1}\left( \max_{j \in \{1, \ldots, k\}} \text{IoU}_{i,j} > 0.5 \right),
\end{equation}
}
where \( N \) represents the total number of negative-positive pairs, and \( \mathbbm{1}(\cdot) \) is an indicator function that equals 1 if the condition inside is true and 0 otherwise. The term \(\text{IoU}_{i,j}\) refers to the overlap between the \(j\)-th predicted bounding box (ranked based on the confidence scores) and the ground truth bounding box for the \(i\)-th pair. Note that for negative samples, there is no ground truth box, meaning the IoU is 0.

Recall@k is commonly used in retrieval tasks to assess prediction accuracy in the presence of challenging negative samples. Ideally, the model should assign lower confidence scores to negative samples. In our study, we primarily report Recall@1. If the model consistently assigns lower confidence scores to negative samples compared to positive ones, Recall@1 should equal Precision@1.

\textbf{AUROC}:  
While Recall@k evaluates how well the model ranks individual negative samples relative to their corresponding positive samples, it does not offer a holistic view of confidence across the dataset. To address this, we use AUROC to measure the overall ability of the model to distinguish between positive and negative samples.  
AUROC measures the model's ability to correctly rank positive samples higher than negative ones across the datasets, providing a holistic view of its discriminative power.

By combining Recall@k and AUROC, we ensure a comprehensive evaluation of the model's performance in distinguishing between positive and negative samples in REC tasks. This dual approach addresses both specific ranking and overall confidence.

\section{Experiment}


\begin{table}[th]
\centering
\resizebox{1\columnwidth}{!}{
\begin{tabular}{lcccc} 
\toprule
                    & \multicolumn{4}{c}{\textbf{Positive}}                                  \\ 
\cmidrule(lr){2-5}
\textbf{Model}      & L1             & L2             & L3             & \textbf{Avg.}            \\ 
\midrule
\textbf{Specialist} &                &                &                &                 \\
Mdetr               & 72.43          & 52.79          & 46.92          & 57.38           \\
MM-GDINO-T          & 75.11          & 34.78          & 35.46          & 48.45           \\
MM-GDINO-L          & 85.13          & 43.54          & 42.89          & 57.19           \\
UNINEXT             & 59.95          & 43.60          & 40.98          & 48.18           \\
MM-GDINO-T$\dagger$          & \uline{85.79}  & 51.88          & 52.65          & 63.44           \\
MM-GDINO-T$\ddagger$        & 82.22          & 51.7           & 51.17          & 61.70           \\ 
\midrule
\textbf{MLLM}       &                &                &                &                 \\
Shikra              & 64.64          & 50.29          & 43.95          & 52.96           \\
Ferret-13B          & 68.24          & 54.88          & 47.56          & 56.89           \\
GroundingGPT        & 71.01          & 53.35          & 49.89          & 58.08           \\
Lenna               & 73.75          & 41.92          & 38.43          & 51.37           \\
InternVL~           & 51.40          & 45.07          & 43.92          & 46.80           \\
CogVLM              & 74.59          & \uline{62.49}  & 57.11          & 64.73           \\
CogCom              & 76.23          & 60.86          & \uline{60.08}  & \uline{65.72}   \\
GPT4-V + SoM        & 55.94          & 45.94          & 49.29          & 50.39           \\
CogVLM$\dagger$               & \textbf{89.23} & \textbf{72.74} & \textbf{72.61} & \textbf{78.19}  \\
\bottomrule
\end{tabular}}
\caption{\label{benchmark_precision}
Evaluation results (Precision@1) on positive data. $\dagger$ indicates training with positive samples from the training set, and $\ddagger$ indicates training with the entire training set. The best results are in bold, and the second-best results are underlined. The same notation will be used in subsequent tables.
}
\end{table}

\subsection{Evaluation settings.}
We evaluates several representative models, including both traditional vision-language models (Specialist) and MLLMs. The models examined in this study include MDETR~\cite{kamath2021mdetr}, MM-GDINO~\cite{zhao2024open,liu2023grounding}, UNINEXT~\cite{yan2023universal}, Shikra~\cite{chen2023shikra}, Ferret~\cite{you2024ferret}, GroundingGPT~\cite{li-etal-2024-groundinggpt}, Lenna~\cite{wei2023lenna}, InternVL~\cite{chen2024far}, CogVLM~\cite{wang2023cogvlm} and CogCom~\cite{qi2024cogcom}. We use there open-source checkpoints to evaluate.
We additionaly evaluate the GPT4-V\cite{achiam2023gpt}.
Since GPT4-V's ability to directly output bounding boxes is relatively limited, we use GPT4-V combined with the Set-of-Mark (SoM)~\cite{yang2023set} to evaluate its performance. The Model source and implementation details are in Appendix~\ref{sec:appendix_exp}.

We also test the effectiveness of training with the training dataset constructed using our data generation pipeline. We fine-tuned MM-GDINO-T and CogVLM using the positive data from the constructed training set. In addition, we fine-tuned MM-GDINO-T with the entire training set. The training settings are detailed in Appendix~\ref{sec:appendix_exp}.

We evaluate the models using Precision@1 for positive data, and Recall@1 and AUROC for negative data; the AUROC results can be found in Appendix~\ref{sec:appendix_results}. Specifically, models like MDETR and Lenna that have dedicated object detection modules can generate multiple detection boxes with associated confidence scores, allowing for direct computation of Recall@1 and AUROC. For models that generate coordinates as text using an autoregressive approach, we use the probability of the coordinate tokens to calculate confidence~\cite{kurita2023refego,mitchell2023detectgpt}.


\begin{table*}[t]
\centering
\resizebox{0.8\textwidth}{!}{
\begin{tabular}{lccccccccccc} 
\toprule
                    & \multicolumn{6}{c}{\textbf{REPLACE}}                                                                                 & \multicolumn{4}{c}{\textbf{SWAP}}                                            &                 \\ 
\cmidrule(lr){2-7}\cmidrule(lr){8-11}
                    & \multicolumn{2}{c}{\textbf{Object}} & \multicolumn{2}{c}{\textbf{Attribute}} & \multicolumn{2}{c}{\textbf{Relation}} & \multicolumn{2}{c}{\textbf{Object}} & \multicolumn{2}{c}{\textbf{Attribute}} &                 \\ 
\cmidrule(lr){2-3}\cmidrule(lr){4-5}\cmidrule(lr){6-7}\cmidrule(r){8-9}\cmidrule(lr){10-11}
\textbf{Model}      & L1             & L2                 & L1             & L2                    & L1             & L2                   & L1             & L2                 & L1             & L2                    & \textbf{Avg.}   \\ 
\midrule
\textbf{Specialist} &                &                    &                &                       &                &                      &                &                    &                &                       &                 \\
MDETR               & 52.89          & 36.09              & 50.47          & 35.92                 & 42.48          & 40.77                & 45.89          & 37.35              & 44.42          & 37.70                 & 42.40           \\
MM-GDINO-T          & 58.84          & 33.77              & 50.47          & 29.96                 & 34.69          & 31.92                & 43.89          & 27.71              & 43.67          & 31.97                 & 38.69           \\
MM-GDINO-L          & 64.23          & 40.26              & 55.76          & 41.52                 & 45.74          & 43.73                & 53.02          & 48.19              & 49.38          & 37.70                 & 47.95           \\
UNINEXT             & 47.83          & 33.70              & 44.66          & 34.30                 & 39.51          & 35.61                & 45.31          & 37.35              & 41.69          & 31.97                 & 39.19           \\
MM-GDINO-T$\dagger$      & \uline{67.60}  & 44.29              & 52.60          & 42.06                 & 48.26          & 46.86                & \uline{59.38}  & 42.77              & \uline{54.34}  & 42.62                 & 50.08           \\
MM-GDINO-T$\ddagger$      & \textbf{72.63} & \textbf{64.87}     & \textbf{68.23} & \textbf{58.84}        & \textbf{62.79} & \textbf{61.07}       & \textbf{65.94} & \textbf{63.25}     & \textbf{68.24} & \textbf{68.03}        & \textbf{65.39}  \\ 
\midrule
\textbf{MLLM}       &                &                    &                &                       &                &                      &                &                    &                &                       &                 \\
Shikra              & 44.99          & 33.11              & 41.25          & 33.03                 & 35.78          & 39.85                & 42.27          & 39.16              & 39.70          & 32.79                 & 38.19           \\
Ferret-13B          & 38.38          & 33.01              & 37.57          & 34.48                 & 35.58          & 34.69                & 38.69          & 34.94              & 35.73          & 35.25                 & 35.83           \\
GroundingGPT        & 42.24          & 35.13              & 40.14          & 33.75                 & 37.51          & 36.72                & 41.77          & 39.76              & 35.24          & 39.34                 & 38.16           \\
Lenna               & 65.88          & \uline{50.38}      & 58.75          & 42.96                 & 47.00          & 43.91                & 49.94          & 38.55              & 49.38          & 43.44                 & 49.02           \\
CogVLM              & 53.34          & 44.02              & 51.24          & 48.74                 & 41.22          & 44.46                & 47.69          & \uline{49.40}      & 46.40          & 40.16                 & 46.67           \\
CogCom              & 57.96          & 44.91              & 54.65          & 44.04                 & 45.81          & 41.70                & 51.03          & 43.98              & 47.39          & 36.89                 & 46.84           \\
CogVLM$\dagger$        & 67.08          & 50.31              & \uline{59.78}  & \uline{53.07}         & \uline{52.78}  & \uline{52.4}         & 53.73          & 49.4               & 52.85          & \uline{50.82}         & \uline{54.22}   \\
\bottomrule
\end{tabular}}
\caption{\label{benchmark_text_recall}
Evaluation results (Recall@1) on negative expressions.
}
\end{table*}

\begin{table*}
\centering
\resizebox{0.8\textwidth}{!}{
\begin{tabular}{lcccccccccc} 
\toprule
                    & \multicolumn{4}{c}{\textbf{REPLACE}}                                         & \multicolumn{5}{c}{\textbf{SWAP}}                                                            &                 \\ 
\cmidrule(lr){2-5}\cmidrule(r){6-10}
                    & \multicolumn{2}{c}{\textbf{Object}} & \multicolumn{2}{c}{\textbf{Attribute}} & \textbf{Object} & \multicolumn{2}{c}{\textbf{Attribute}} & \multicolumn{2}{c}{\textbf{Flip}} &                 \\ 
\cmidrule(r){2-3}\cmidrule(r){4-5}\cmidrule(lr){6-6}\cmidrule(r){7-8}\cmidrule(lr){9-10}
\textbf{Model}      & L1             & L2                 & L1             & L2                    & L1              & L1             & L2                    & L1             & L2               & \textbf{Avg.}   \\ 
\midrule
\textbf{Specialist} &                &                    &                &                       &                 &                &                       &                &                  &                 \\
MDETR               & 58.15          & 42.85              & 51.70          & 37.95                 & 48.86           & 49.49          & 44.76                 & 44.29          & 42.22            & 46.70           \\
MM-GDINO-T          & 58.46          & 40.73              & 44.75          & 37.61                 & 46.25           & 51.33          & 28.67                 & 39.50          & 40.94            & 43.14           \\
MM-GDINO-L          & 66.35          & 49.45              & 54.93          & 49.05                 & 55.05           & 62.63          & 46.85                 & 45.21          & 46.48            & 52.89           \\
UNINEXT             & 48.85          & 31.62              & 40.96          & 30.33                 & 46.91           & 40.25          & 37.06                 & 30.66          & 29.42            & 37.34           \\
MM-GDINO-T$\dagger$      & \uline{70.37}  & 55.68              & \uline{56.83}  & 53.73                 & \textbf{57.98}  & \uline{62.83}  & 55.24                 & \uline{48.71}  & \textbf{52.03}   & \uline{57.04}   \\
MM-GDINO-T$\ddagger$      & \textbf{74.46} & \textbf{64.59}     & \textbf{65.35} & \textbf{63.43}        & 55.70           & \textbf{67.97} & \textbf{72.73}        & 45.86          & 47.55            & \textbf{61.96}  \\ 
\midrule
\textbf{MLLM}       &                &                    &                &                       &                 &                &                       &                &                  &                 \\
Shikra              & 42.57          & 33.61              & 36.54          & 34.26                 & 35.18           & 38.60          & 36.36                 & 34.25          & 37.10            & 36.50           \\
Ferret-13B          & 41.54          & 37.46              & 38.04          & 36.22                 & 43.00           & 37.78          & 39.16                 & 35.27          & 36.25            & 38.30           \\
GroundingGPT        & 43.91          & 36.88              & 36.31          & 35.88                 & 39.09           & 37.17          & 40.56                 & 37.02          & 33.05            & 37.76           \\
Lenna               & 66.88          & 51.19              & 54.38          & 39.34                 & 47.56           & 49.08          & 43.36                 & 33.98          & 30.92            & 46.30           \\
CogVLM              & 51.11          & 49.01              & 43.49          & 46.10                 & 50.49           & 53.80          & 49.65                 & 43.74          & 37.74            & 47.24           \\
CogCom              & 32.24          & 21.55              & 22.57          & 20.10                 & 39.74           & 25.46          & 18.88                 & 24.13          & 23.03            & 25.30           \\
CogVLM$\dagger$        & 62.02          & \uline{55.81}      & 46.41          & \uline{55.98}         & \uline{56.35}   & 55.03          & \uline{57.34}         & \textbf{49.08} & \uline{48.83}    & 54.09           \\
\bottomrule
\end{tabular}}
\caption{\label{benchmark_img_recall}
Evaluation results (Recall@1) on negative images. 
}
\end{table*}

\subsection{Evaluation on Positive data}
The results shown in Table~\ref{benchmark_precision} indicate that categorizing the dataset by difficulty level is crucial, as the performance of the most of the models declines with increasing difficulty. Notably, for level 3, most models achieve a precision below 50\%.

\textbf{Specialist perform better on simple REC task.}
At level 1, models merely need to detect objects based on their names, aligning with the requirements of open-vocabulary object detection. It was observed that Grounding DINO, based on SWIN-L, achieved an accuracy of 85.13\% under zero-shot settings. This leads to two conclusions. First, vision-language models focused on object detection exhibit strong capabilities in basic visual localization and object detection tasks, even in zero-shot scenarios, which is also supported by their superior performance on RefCOCO benchmark which mainly require the model to detect the obejct without consider the attribute and relation. Second, although multimodal large models excel in dialogue and language understanding, their basic object detection abilities still fall short of the standards required for truly general-purpose models.

\textbf{MLLMs exhibit superior reasoning abilities.}
For levels 2 and 3, models need robust language comprehension due to the presence of many easily confusable objects in the images. However, most models do not demonstrate sufficient capability in this aspect.
Multimodal models based on large language models (LLMs) achieved better results in this regard, demonstrating that MLLMs possess stronger compositional reasoning abilities.

\begin{figure*}[t]
\centering
  \includegraphics[width=0.9\textwidth]{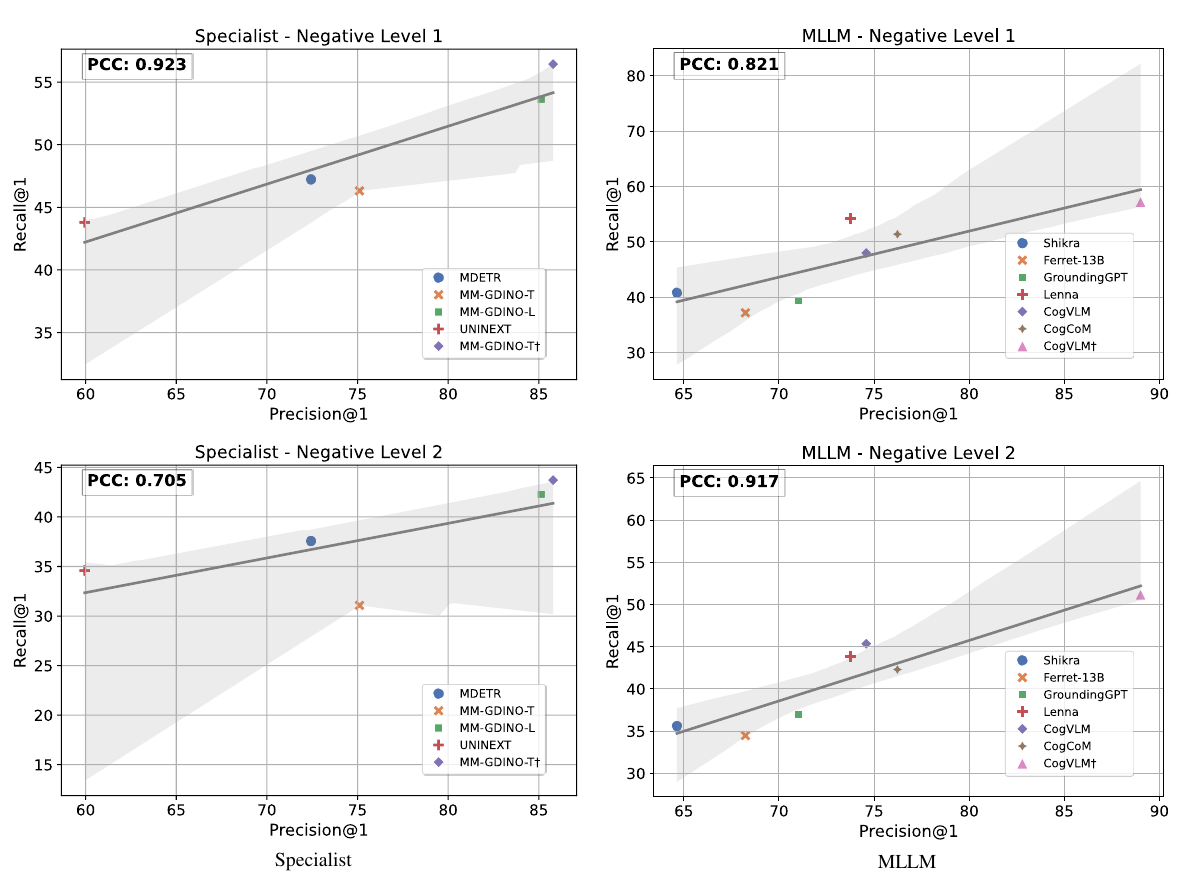}
  \caption{The relationship between Precision@1 (on positive samples) and Recall@1 (on positive and negative samples) for Specialist and MLLM models across different negative difficulty levels. Specialist models correlate strongly with easier negative samples (Negative level 1, PCC = 0.923), while MLLMs show a higher correlation with harder negatives (Negative level 2, PCC = 0.917), reflecting their differing focuses on compositional REC.}
  \label{fig:precision_recall}
\end{figure*}

\subsection{Evaluation on Negative data}
The evaluation results for negative expressions and negative images are shown in Table~\ref{benchmark_text_recall} and Table~\ref{benchmark_img_recall}, respectively. We can draw the following conclusions:

\textbf{The models are highly sensitive to the specific locations of negative data.}
L1 and L2 represent the replacement of the target directly and the replacement of other parts of the expression, respectively.
For most types of negative data, the recall for L1 is significantly higher than for L2. This indicates that most models can identify simple anomalies, such as changes in the main target or inconsistencies in relationships.
However, for L2 negative data, all models perform poorly, further demonstrating that the models lack compositional reasoning abilities and do not pay attention to the complete structure of the sentences.

\textbf{The models have poor understanding of relationships.}
Overall, the models show relatively good recognition capabilities for direct object replacements, where the target mentioned in the expression is entirely absent in the image. Their ability to recognize attributes is slightly weaker.
The models struggle significantly with understanding relationships, including recognizing replaced relationships and altered word order, which aligns with findings from previous studies.
An additional finding is that the models perform worse in recognizing negative data of the "swap attribute" type compared to direct attribute replacements, indicating limitations in the models' ability to bind attributes accurately.

\section{In depth analysis}

\subsection{What's the relationship between Precision and Recall?}
In Figure~\ref{fig:precision_recall}, we explored the relationship between Precision@1 and Recall@1 among models. It is clearly evident that \textbf{Precision and Recall are positively correlated}. This is consistent with the findings of \citet{ma2023crepe, vaze2022openset}, where the accuracy of models on positive samples typically correlates positively with their ability to identify or reject out-of-distribution (OOD) samples.

Additionally, we further analyzed the correlation of different model types with different levels of negatives. We discovered a particularly interesting phenomenon: the precision of Specialist models has a Pearson Correlation Coefficient (PCC) of 0.923 with Negative level 1, whereas the precision of MLLMs has a PCC of 0.917 with Negative level 2. This further confirms the differing tendencies of MLLM and Specialist models. Specifically, \textbf{Specialist tend to learn the existence and attributes of targets, while MLLMs models focus more on compositional reasoning.}

\subsection{Is rewrite useful?}

To verify the significance of rewriting benchmark data, we conducted comparative experiments where models were evaluated using both the original data and the rewritten data. As shown in Table~\ref{rewrite_template}, models achieved significantly better performance on the evaluation benchmark without rewriting.

For positive data, using template-generated data always places the subject at the beginning of the sentence and has a very clear linguistic structure, which does not adequately assess the model's language understanding abilities. For negative data, without rewriting, there are issues with non-fluency and nonsensicality~\cite{hsieh_sugarcrepe_2023}, which can be easily detected by text-only models such as Grammar~\cite{morris-etal-2020-textattack}.

\begin{table}[t]
\centering
\resizebox{1\columnwidth}{!}{
\begin{tabular}{lccc} 
\toprule
\textbf{Model}      & Rewrite & Percision@1 & Recall@1  \\ 
\midrule
MM-GDINO-T & \ding{55}      & 50.23       & 44.42     \\
MM-GDINO-T & \ding{51}     & 48.45       & 38.69
    \\ 
\midrule
CogVLM     & \ding{55}      & 71.18       & 52.34     \\
CogVLM     & \ding{51}     & 64.73       & 46.67
    \\ 
\midrule
Grammar    & \ding{55}      & -       & 54.63     \\
Grammar    & \ding{51}     & -       & 50.21     \\ 
\bottomrule
\end{tabular}}
\caption{\label{rewrite_template}
Ablation study on the effect of rewriting the benchmark dataset. The reported metrics are the average Precision@1 and Recall@1 scores.
}
\end{table}

\begin{table}[t]
\centering
\resizebox{1\columnwidth}{!}{
\begin{tabular}{lcccccccc} 
\toprule
               & \multicolumn{3}{c}{\textbf{RefCOCO}}             & \multicolumn{3}{c}{\textbf{RefCOCO+}}            & \multicolumn{2}{c}{\textbf{RefCOCOg}}  \\ 
\cmidrule(lr){2-4}\cmidrule(r){5-7}\cmidrule(lr){8-9}
\textbf{Model} & val            & test-A         & test-B         & val            & test-A         & test-B         & val            & test                  \\ 
\midrule
CogVLM         & 92.76          & 94.75          & 88.99          & 88.68          & 92.91          & 83.39          & 89.75          & 90.79                 \\
CogVLM$\dagger$         & \textbf{93.11} & \textbf{95.02} & \textbf{89.95} & \textbf{88.72} & \textbf{92.94} & \textbf{83.50} & \textbf{90.75} & \textbf{91.19}        \\
\bottomrule
\end{tabular}}
\caption{\label{benchmark_refcoco_precision}
Evaluation results (Precision@1) on RefCOCO/+/g. The results of CogVLM come from the original paper.}
\end{table}

\subsection{Evaluation on RefCOCO}

We additionally validated the performance of the CogVLM fine-tuned on our training set with RefCOCO/+/g benchmarks. As shown in Table~\ref{benchmark_refcoco_precision}, our model outperformed the original CogVLM in all validation and test sets. This result demonstrates the high quality and generalization capabilities of our dataset.

\section{Conclusion}

In this work, we introduced FineCops-Ref, a novel dataset for fine-grained compositional referring expression comprehension with varying difficulty levels and negative samples. Our evaluations reveal that while current MLLMs perform well on traditional REC benchmarks, they struggle with advanced compositional reasoning and accurate rejection of negative samples. 
Our dataset provides strong support for the evaluation of the model's compositional grounding ability, and the training set can also serve as a good supplement to existing training data for compositional REC. 
We hope FineCops-Ref can inspire further research into enhancing compositional visual grounding.


\section{Limitations}
We employ LLMs and diffusion models for data generation, which inevitably introduce some hallucinations. Despite manual filtering of the benchmark dataset, hallucinations still persist in the training set. Additionally, while the models fine-tuned on the proposed training set exhibit good performance, we still lack effective methods for effectively recognizing hallucinations and handling negative samples.

Furthermore, although REC can evaluate the grounding ability of the model, the relationship between performance on REC tasks and other tasks such as VQA still needs to be explored. We also lack a complete evaluation of the model's conversational abilities, like grounded image captions.



\bibliography{anthology,custom}

\begin{thebibliography}{60}
\providecommand{\natexlab}[1]{#1}

\bibitem[{Achiam et~al.(2023)Achiam, Adler, Agarwal, Ahmad, Akkaya, Aleman, Almeida, Altenschmidt, Altman, Anadkat et~al.}]{achiam2023gpt}
Josh Achiam, Steven Adler, Sandhini Agarwal, Lama Ahmad, Ilge Akkaya, Florencia~Leoni Aleman, Diogo Almeida, Janko Altenschmidt, Sam Altman, Shyamal Anadkat, et~al. 2023.
\newblock Gpt-4 technical report.
\newblock \emph{arXiv preprint arXiv:2303.08774}.

\bibitem[{Akula et~al.(2020)Akula, Gella, Al-Onaizan, Zhu, and Reddy}]{akula-etal-2020-words}
Arjun Akula, Spandana Gella, Yaser Al-Onaizan, Song-Chun Zhu, and Siva Reddy. 2020.
\newblock \href {https://doi.org/10.18653/v1/2020.acl-main.586} {Words aren`t enough, their order matters: On the robustness of grounding visual referring expressions}.
\newblock In \emph{Proceedings of the 58th Annual Meeting of the Association for Computational Linguistics}, pages 6555--6565, Online. Association for Computational Linguistics.

\bibitem[{Chen et~al.(2024{\natexlab{a}})Chen, Xu, Kirmani, Ichter, Driess, Florence, Sadigh, Guibas, and Xia}]{chen2024spatialvlm}
Boyuan Chen, Zhuo Xu, Sean Kirmani, Brian Ichter, Danny Driess, Pete Florence, Dorsa Sadigh, Leonidas Guibas, and Fei Xia. 2024{\natexlab{a}}.
\newblock \href {https://arxiv.org/abs/2401.12168} {Spatialvlm: Endowing vision-language models with spatial reasoning capabilities}.
\newblock \emph{arXiv preprint arXiv:2401.12168}.

\bibitem[{Chen et~al.(2023)Chen, Zhang, Zeng, Zhang, Zhu, and Zhao}]{chen2023shikra}
Keqin Chen, Zhao Zhang, Weili Zeng, Richong Zhang, Feng Zhu, and Rui Zhao. 2023.
\newblock Shikra: Unleashing multimodal llm's referential dialogue magic.
\newblock \emph{arXiv preprint arXiv:2306.15195}.

\bibitem[{Chen et~al.(2024{\natexlab{b}})Chen, Wang, Tian, Ye, Gao, Cui, Tong, Hu, Luo, Ma et~al.}]{chen2024far}
Zhe Chen, Weiyun Wang, Hao Tian, Shenglong Ye, Zhangwei Gao, Erfei Cui, Wenwen Tong, Kongzhi Hu, Jiapeng Luo, Zheng Ma, et~al. 2024{\natexlab{b}}.
\newblock How far are we to gpt-4v? closing the gap to commercial multimodal models with open-source suites.
\newblock \emph{arXiv preprint arXiv:2404.16821}.

\bibitem[{Chen et~al.(2020)Chen, Wang, Ma, Wong, and Wu}]{chen2020cops}
Zhenfang Chen, Peng Wang, Lin Ma, Kwan-Yee~K Wong, and Qi~Wu. 2020.
\newblock Cops-ref: A new dataset and task on compositional referring expression comprehension.
\newblock In \emph{Proceedings of the IEEE/CVF Conference on Computer Vision and Pattern Recognition}, pages 10086--10095.

\bibitem[{Cirik et~al.(2018)Cirik, Morency, and Berg-Kirkpatrick}]{cirik-etal-2018-visual}
Volkan Cirik, Louis-Philippe Morency, and Taylor Berg-Kirkpatrick. 2018.
\newblock \href {https://doi.org/10.18653/v1/N18-2123} {Visual referring expression recognition: What do systems actually learn?}
\newblock In \emph{Proceedings of the 2018 Conference of the North {A}merican Chapter of the Association for Computational Linguistics: Human Language Technologies, Volume 2 (Short Papers)}, pages 781--787, New Orleans, Louisiana. Association for Computational Linguistics.

\bibitem[{Diwan et~al.(2022)Diwan, Berry, Choi, Harwath, and Mahowald}]{diwan-etal-2022-winoground}
Anuj Diwan, Layne Berry, Eunsol Choi, David Harwath, and Kyle Mahowald. 2022.
\newblock \href {https://doi.org/10.18653/v1/2022.emnlp-main.143} {Why is winoground hard? investigating failures in visuolinguistic compositionality}.
\newblock In \emph{Proceedings of the 2022 Conference on Empirical Methods in Natural Language Processing}, pages 2236--2250, Abu Dhabi, United Arab Emirates. Association for Computational Linguistics.

\bibitem[{Hsieh et~al.(2023)Hsieh, Zhang, Ma, Kembhavi, and Krishna}]{hsieh_sugarcrepe_2023}
Cheng-Yu Hsieh, Jieyu Zhang, Zixian Ma, Aniruddha Kembhavi, and Ranjay Krishna. 2023.
\newblock \href {https://proceedings.neurips.cc/paper_files/paper/2023/file/63461de0b4cb760fc498e85b18a7fe81-Paper-Datasets_and_Benchmarks.pdf} {{SugarCrepe}: {Fixing} {Hackable} {Benchmarks} for {Vision}-{Language} {Compositionality}}.
\newblock In \emph{Advances in {Neural} {Information} {Processing} {Systems}}, volume~36, pages 31096--31116. Curran Associates, Inc.

\bibitem[{Hu et~al.(2022)Hu, yelong shen, Wallis, Allen-Zhu, Li, Wang, Wang, and Chen}]{hu2022lora}
Edward~J Hu, yelong shen, Phillip Wallis, Zeyuan Allen-Zhu, Yuanzhi Li, Shean Wang, Lu~Wang, and Weizhu Chen. 2022.
\newblock \href {https://openreview.net/forum?id=nZeVKeeFYf9} {Lo{RA}: Low-rank adaptation of large language models}.
\newblock In \emph{International Conference on Learning Representations}.

\bibitem[{Hudson and Manning(2019)}]{hudson2019gqa}
Drew~A Hudson and Christopher~D Manning. 2019.
\newblock Gqa: A new dataset for real-world visual reasoning and compositional question answering.
\newblock In \emph{Proceedings of the IEEE/CVF conference on computer vision and pattern recognition}, pages 6700--6709.

\bibitem[{Kamath et~al.(2021)Kamath, Singh, LeCun, Synnaeve, Misra, and Carion}]{kamath2021mdetr}
Aishwarya Kamath, Mannat Singh, Yann LeCun, Gabriel Synnaeve, Ishan Misra, and Nicolas Carion. 2021.
\newblock Mdetr-modulated detection for end-to-end multi-modal understanding.
\newblock In \emph{Proceedings of the IEEE/CVF International Conference on Computer Vision}, pages 1780--1790.

\bibitem[{Kurita et~al.(2023)Kurita, Katsura, and Onami}]{kurita2023refego}
Shuhei Kurita, Naoki Katsura, and Eri Onami. 2023.
\newblock Refego: Referring expression comprehension dataset from first-person perception of ego4d.
\newblock In \emph{Proceedings of the IEEE/CVF International Conference on Computer Vision}, pages 15214--15224.

\bibitem[{Le et~al.(2023)Le, LAL, and Howard}]{NEURIPS2023_coco_counter}
Tiep Le, VASUDEV LAL, and Phillip Howard. 2023.
\newblock \href {https://proceedings.neurips.cc/paper_files/paper/2023/file/e14e4cb8266184ceb234973dfe07faed-Paper-Datasets_and_Benchmarks.pdf} {Coco-counterfactuals: Automatically constructed counterfactual examples for image-text pairs}.
\newblock In \emph{Advances in Neural Information Processing Systems}, volume~36, pages 71195--71221. Curran Associates, Inc.

\bibitem[{Li et~al.(2024{\natexlab{a}})Li, Chen, Hong, Chen, Chen, Shen, and Gan}]{li2024covlm}
Junyan Li, Delin Chen, Yining Hong, Zhenfang Chen, Peihao Chen, Yikang Shen, and Chuang Gan. 2024{\natexlab{a}}.
\newblock \href {https://openreview.net/forum?id=PHGxChm1l5} {Co{VLM}: Composing visual entities and relationships in large language models via communicative decoding}.
\newblock In \emph{The Twelfth International Conference on Learning Representations}.

\bibitem[{Li et~al.(2024{\natexlab{b}})Li, Xu, Zhang, Song, Cai, Qi, Zhou, Pan, Li, Tu, Huang, and Wang}]{li-etal-2024-groundinggpt}
Zhaowei Li, Qi~Xu, Dong Zhang, Hang Song, YiQing Cai, Qi~Qi, Ran Zhou, Junting Pan, Zefeng Li, Vu~Tu, Zhida Huang, and Tao Wang. 2024{\natexlab{b}}.
\newblock \href {https://doi.org/10.18653/v1/2024.acl-long.360} {{G}rounding{GPT}: Language enhanced multi-modal grounding model}.
\newblock In \emph{Proceedings of the 62nd Annual Meeting of the Association for Computational Linguistics (Volume 1: Long Papers)}, pages 6657--6678, Bangkok, Thailand. Association for Computational Linguistics.

\bibitem[{Lin et~al.(2023)Lin, Chen, Pathak, Zhang, and Ramanan}]{lin2023visualgptscore}
Zhiqiu Lin, Xinyue Chen, Deepak Pathak, Pengchuan Zhang, and Deva Ramanan. 2023.
\newblock Visualgptscore: Visio-linguistic reasoning with multimodal generative pre-training scores.
\newblock \emph{arXiv preprint arXiv:2306.01879}.

\bibitem[{Liu et~al.(2023{\natexlab{a}})Liu, Emerson, and Collier}]{liu-etal-2023-visual}
Fangyu Liu, Guy Emerson, and Nigel Collier. 2023{\natexlab{a}}.
\newblock \href {https://doi.org/10.1162/tacl_a_00566} {Visual spatial reasoning}.
\newblock \emph{Transactions of the Association for Computational Linguistics}, 11:635--651.

\bibitem[{Liu et~al.(2019)Liu, Liu, Bai, and Yuille}]{liu2019clevr}
Runtao Liu, Chenxi Liu, Yutong Bai, and Alan~L Yuille. 2019.
\newblock Clevr-ref+: Diagnosing visual reasoning with referring expressions.
\newblock In \emph{Proceedings of the IEEE/CVF conference on computer vision and pattern recognition}, pages 4185--4194.

\bibitem[{Liu et~al.(2023{\natexlab{b}})Liu, Zeng, Ren, Li, Zhang, Yang, Li, Yang, Su, Zhu et~al.}]{liu2023grounding}
Shilong Liu, Zhaoyang Zeng, Tianhe Ren, Feng Li, Hao Zhang, Jie Yang, Chunyuan Li, Jianwei Yang, Hang Su, Jun Zhu, et~al. 2023{\natexlab{b}}.
\newblock Grounding dino: Marrying dino with grounded pre-training for open-set object detection.
\newblock \emph{arXiv preprint arXiv:2303.05499}.

\bibitem[{Ma et~al.(2024)Ma, Jiang, Wu, Yuan, and Qi}]{ma2024groma}
Chuofan Ma, Yi~Jiang, Jiannan Wu, Zehuan Yuan, and Xiaojuan Qi. 2024.
\newblock Groma: Localized visual tokenization for grounding multimodal large language models.
\newblock \emph{arXiv preprint arXiv:2404.13013}.

\bibitem[{Ma et~al.(2023)Ma, Hong, Gul, Gandhi, Gao, and Krishna}]{ma2023crepe}
Zixian Ma, Jerry Hong, Mustafa~Omer Gul, Mona Gandhi, Irena Gao, and Ranjay Krishna. 2023.
\newblock Crepe: Can vision-language foundation models reason compositionally?
\newblock In \emph{Proceedings of the IEEE/CVF Conference on Computer Vision and Pattern Recognition}, pages 10910--10921.

\bibitem[{Mitchell et~al.(2023)Mitchell, Lee, Khazatsky, Manning, and Finn}]{mitchell2023detectgpt}
Eric Mitchell, Yoonho Lee, Alexander Khazatsky, Christopher~D Manning, and Chelsea Finn. 2023.
\newblock Detectgpt: Zero-shot machine-generated text detection using probability curvature.
\newblock In \emph{International Conference on Machine Learning}, pages 24950--24962. PMLR.

\bibitem[{Miyai et~al.(2024)Miyai, Yang, Zhang, Ming, Yu, Irie, Li, Li, Liu, and Aizawa}]{miyai2024unsolvable}
Atsuyuki Miyai, Jingkang Yang, Jingyang Zhang, Yifei Ming, Qing Yu, Go~Irie, Yixuan Li, Hai Li, Ziwei Liu, and Kiyoharu Aizawa. 2024.
\newblock Unsolvable problem detection: Evaluating trustworthiness of vision language models.
\newblock \emph{arXiv preprint arXiv:2403.20331}.

\bibitem[{Morris et~al.(2020)Morris, Lifland, Yoo, Grigsby, Jin, and Qi}]{morris-etal-2020-textattack}
John Morris, Eli Lifland, Jin~Yong Yoo, Jake Grigsby, Di~Jin, and Yanjun Qi. 2020.
\newblock \href {https://doi.org/10.18653/v1/2020.emnlp-demos.16} {{T}ext{A}ttack: A framework for adversarial attacks, data augmentation, and adversarial training in {NLP}}.
\newblock In \emph{Proceedings of the 2020 Conference on Empirical Methods in Natural Language Processing: System Demonstrations}, pages 119--126, Online. Association for Computational Linguistics.

\bibitem[{Oquab et~al.(2023)Oquab, Darcet, Moutakanni, Vo, Szafraniec, Khalidov, Fernandez, Haziza, Massa, El-Nouby et~al.}]{oquab2023dinov2}
Maxime Oquab, Timoth{\'e}e Darcet, Th{\'e}o Moutakanni, Huy Vo, Marc Szafraniec, Vasil Khalidov, Pierre Fernandez, Daniel Haziza, Francisco Massa, Alaaeldin El-Nouby, et~al. 2023.
\newblock Dinov2: Learning robust visual features without supervision.
\newblock \emph{arXiv preprint arXiv:2304.07193}.

\bibitem[{Parcalabescu et~al.(2022)Parcalabescu, Cafagna, Muradjan, Frank, Calixto, and Gatt}]{parcalabescu-etal-2022-valse}
Letitia Parcalabescu, Michele Cafagna, Lilitta Muradjan, Anette Frank, Iacer Calixto, and Albert Gatt. 2022.
\newblock \href {https://doi.org/10.18653/v1/2022.acl-long.567} {{VALSE}: A task-independent benchmark for vision and language models centered on linguistic phenomena}.
\newblock In \emph{Proceedings of the 60th Annual Meeting of the Association for Computational Linguistics (Volume 1: Long Papers)}, pages 8253--8280, Dublin, Ireland. Association for Computational Linguistics.

\bibitem[{Peng et~al.(2024)Peng, Wang, Dong, Hao, Huang, Ma, Ye, and Wei}]{peng2024grounding}
Zhiliang Peng, Wenhui Wang, Li~Dong, Yaru Hao, Shaohan Huang, Shuming Ma, Qixiang Ye, and Furu Wei. 2024.
\newblock \href {https://openreview.net/forum?id=lLmqxkfSIw} {Grounding multimodal large language models to the world}.
\newblock In \emph{The Twelfth International Conference on Learning Representations}.

\bibitem[{Pramanick et~al.(2023)Pramanick, Han, Hou, Nag, Lim, Ballas, Wang, Chellappa, and Almahairi}]{pramanick2023jack}
Shraman Pramanick, Guangxing Han, Rui Hou, Sayan Nag, Ser-Nam Lim, Nicolas Ballas, Qifan Wang, Rama Chellappa, and Amjad Almahairi. 2023.
\newblock Jack of all tasks, master of many: Designing general-purpose coarse-to-fine vision-language model.
\newblock \emph{arXiv preprint arXiv:2312.12423}.

\bibitem[{Qi et~al.(2024)Qi, Ding, Wang, Bai, Lv, Hong, Xu, Hou, Li, Dong et~al.}]{qi2024cogcom}
Ji~Qi, Ming Ding, Weihan Wang, Yushi Bai, Qingsong Lv, Wenyi Hong, Bin Xu, Lei Hou, Juanzi Li, Yuxiao Dong, et~al. 2024.
\newblock Cogcom: Train large vision-language models diving into details through chain of manipulations.
\newblock \emph{arXiv preprint arXiv:2402.04236}.

\bibitem[{Radford et~al.(2021)Radford, Kim, Hallacy, Ramesh, Goh, Agarwal, Sastry, Askell, Mishkin, Clark et~al.}]{radford2021learning}
Alec Radford, Jong~Wook Kim, Chris Hallacy, Aditya Ramesh, Gabriel Goh, Sandhini Agarwal, Girish Sastry, Amanda Askell, Pamela Mishkin, Jack Clark, et~al. 2021.
\newblock Learning transferable visual models from natural language supervision.
\newblock In \emph{International conference on machine learning}, pages 8748--8763. PMLR.

\bibitem[{Rasheed et~al.(2024)Rasheed, Maaz, Shaji, Shaker, Khan, Cholakkal, Anwer, Xing, Yang, and Khan}]{Rasheed_2024_CVPR}
Hanoona Rasheed, Muhammad Maaz, Sahal Shaji, Abdelrahman Shaker, Salman Khan, Hisham Cholakkal, Rao~M. Anwer, Eric Xing, Ming-Hsuan Yang, and Fahad~S. Khan. 2024.
\newblock Glamm: Pixel grounding large multimodal model.
\newblock In \emph{Proceedings of the IEEE/CVF Conference on Computer Vision and Pattern Recognition (CVPR)}, pages 13009--13018.

\bibitem[{Ray et~al.(2023)Ray, Radenovic, Dubey, Plummer, Krishna, and Saenko}]{ray_cola_2023}
Arijit Ray, Filip Radenovic, Abhimanyu Dubey, Bryan Plummer, Ranjay Krishna, and Kate Saenko. 2023.
\newblock \href {https://proceedings.neurips.cc/paper_files/paper/2023/file/917cd410aa55b61594fa2a6f6e5a9e94-Paper-Datasets_and_Benchmarks.pdf} {Cola: {A} {Benchmark} for {Compositional} {Text}-to-image {Retrieval}}.
\newblock In \emph{Advances in {Neural} {Information} {Processing} {Systems}}, volume~36, pages 46433--46445. Curran Associates, Inc.

\bibitem[{Schulter et~al.(2023)Schulter, Suh, Dafnis, Zhang, Zhao, Metaxas et~al.}]{schulter2023omnilabel}
Samuel Schulter, Yumin Suh, Konstantinos~M Dafnis, Zhixing Zhang, Shiyu Zhao, Dimitris Metaxas, et~al. 2023.
\newblock Omnilabel: A challenging benchmark for language-based object detection.
\newblock In \emph{Proceedings of the IEEE/CVF International Conference on Computer Vision}, pages 11953--11962.

\bibitem[{Shao et~al.(2024)Shao, Qian, Xiao, Song, Zong, Wang, Liu, and Li}]{shao2024visualcot}
Hao Shao, Shengju Qian, Han Xiao, Guanglu Song, Zhuofan Zong, Letian Wang, Yu~Liu, and Hongsheng Li. 2024.
\newblock Visual cot: Unleashing chain-of-thought reasoning in multi-modal language models.
\newblock \emph{arXiv preprint arXiv:2403.16999}.

\bibitem[{Suhr et~al.(2019)Suhr, Zhou, Zhang, Zhang, Bai, and Artzi}]{suhr-etal-2019-corpus}
Alane Suhr, Stephanie Zhou, Ally Zhang, Iris Zhang, Huajun Bai, and Yoav Artzi. 2019.
\newblock \href {https://doi.org/10.18653/v1/P19-1644} {A corpus for reasoning about natural language grounded in photographs}.
\newblock In \emph{Proceedings of the 57th Annual Meeting of the Association for Computational Linguistics}, pages 6418--6428, Florence, Italy. Association for Computational Linguistics.

\bibitem[{Thrush et~al.(2022)Thrush, Jiang, Bartolo, Singh, Williams, Kiela, and Ross}]{thrush2022winoground}
Tristan Thrush, Ryan Jiang, Max Bartolo, Amanpreet Singh, Adina Williams, Douwe Kiela, and Candace Ross. 2022.
\newblock Winoground: Probing vision and language models for visio-linguistic compositionality.
\newblock In \emph{Proceedings of the IEEE/CVF Conference on Computer Vision and Pattern Recognition}, pages 5238--5248.

\bibitem[{Tong et~al.(2024)Tong, Liu, Zhai, Ma, LeCun, and Xie}]{tong2024eyes}
Shengbang Tong, Zhuang Liu, Yuexiang Zhai, Yi~Ma, Yann LeCun, and Saining Xie. 2024.
\newblock Eyes wide shut? exploring the visual shortcomings of multimodal llms.
\newblock In \emph{Proceedings of the IEEE/CVF Conference on Computer Vision and Pattern Recognition (CVPR)}, pages 9568--9578.

\bibitem[{Vaze et~al.(2022)Vaze, Han, Vedaldi, and Zisserman}]{vaze2022openset}
Sagar Vaze, Kai Han, Andrea Vedaldi, and Andrew Zisserman. 2022.
\newblock \href {https://openreview.net/forum?id=5hLP5JY9S2d} {Open-set recognition: A good closed-set classifier is all you need}.
\newblock In \emph{International Conference on Learning Representations}.

\bibitem[{Wang et~al.(2023{\natexlab{a}})Wang, Lv, Yu, Hong, Qi, Wang, Ji, Yang, Zhao, Song et~al.}]{wang2023cogvlm}
Weihan Wang, Qingsong Lv, Wenmeng Yu, Wenyi Hong, Ji~Qi, Yan Wang, Junhui Ji, Zhuoyi Yang, Lei Zhao, Xixuan Song, et~al. 2023{\natexlab{a}}.
\newblock Cogvlm: Visual expert for pretrained language models.
\newblock \emph{arXiv preprint arXiv:2311.03079}.

\bibitem[{Wang et~al.(2024)Wang, Shi, Li, Wang, Huang, Xing, Chen, Li, Zhu, Cao, Chen, Lu, Dai, and Qiao}]{wang2024the}
Weiyun Wang, Min Shi, Qingyun Li, Wenhai Wang, Zhenhang Huang, Linjie Xing, Zhe Chen, Hao Li, Xizhou Zhu, Zhiguo Cao, Yushi Chen, Tong Lu, Jifeng Dai, and Yu~Qiao. 2024.
\newblock \href {https://openreview.net/forum?id=c2R7ajodcI} {The all-seeing project: Towards panoptic visual recognition and understanding of the open world}.
\newblock In \emph{The Twelfth International Conference on Learning Representations}.

\bibitem[{Wang et~al.(2023{\natexlab{b}})Wang, Chen, Chen, Wu, Zhu, Zeng, Luo, Lu, Zhou, Qiao, and Dai}]{wang_visionllm_2023}
Wenhai Wang, Zhe Chen, Xiaokang Chen, Jiannan Wu, Xizhou Zhu, Gang Zeng, Ping Luo, Tong Lu, Jie Zhou, Yu~Qiao, and Jifeng Dai. 2023{\natexlab{b}}.
\newblock \href {https://proceedings.neurips.cc/paper_files/paper/2023/file/c1f7b1ed763e9c75e4db74b49b76db5f-Paper-Conference.pdf} {{VisionLLM}: {Large} {Language} {Model} is also an {Open}-{Ended} {Decoder} for {Vision}-{Centric} {Tasks}}.
\newblock In \emph{Advances in {Neural} {Information} {Processing} {Systems}}, volume~36, pages 61501--61513. Curran Associates, Inc.

\bibitem[{Wang et~al.(2023{\natexlab{c}})Wang, Bao, Zhou, Wang, Hu, Chen, and Li}]{wang2023dire}
Zhendong Wang, Jianmin Bao, Wengang Zhou, Weilun Wang, Hezhen Hu, Hong Chen, and Houqiang Li. 2023{\natexlab{c}}.
\newblock Dire for diffusion-generated image detection.
\newblock In \emph{Proceedings of the IEEE/CVF International Conference on Computer Vision}, pages 22445--22455.

\bibitem[{Wei et~al.(2023)Wei, Zhang, Zhang, Zhang, and Chu}]{wei2023lenna}
Fei Wei, Xinyu Zhang, Ailing Zhang, Bo~Zhang, and Xiangxiang Chu. 2023.
\newblock Lenna: Language enhanced reasoning detection assistant.
\newblock \emph{arXiv preprint arXiv:2312.02433}.

\bibitem[{Wu et~al.(2023)Wu, Wei, Wang, Liu, Yang, and He}]{Wu_2023_ICCV_GITM-MR}
Yu~Wu, Yana Wei, Haozhe Wang, Yongfei Liu, Sibei Yang, and Xuming He. 2023.
\newblock Grounded image text matching with mismatched relation reasoning.
\newblock In \emph{Proceedings of the IEEE/CVF International Conference on Computer Vision (ICCV)}, pages 2976--2987.

\bibitem[{Yan et~al.(2023)Yan, Jiang, Wu, Wang, Luo, Yuan, and Lu}]{yan2023universal}
Bin Yan, Yi~Jiang, Jiannan Wu, Dong Wang, Ping Luo, Zehuan Yuan, and Huchuan Lu. 2023.
\newblock Universal instance perception as object discovery and retrieval.
\newblock In \emph{Proceedings of the IEEE/CVF Conference on Computer Vision and Pattern Recognition}, pages 15325--15336.

\bibitem[{Yang et~al.(2023)Yang, Zhang, Li, Zou, Li, and Gao}]{yang2023set}
Jianwei Yang, Hao Zhang, Feng Li, Xueyan Zou, Chunyuan Li, and Jianfeng Gao. 2023.
\newblock Set-of-mark prompting unleashes extraordinary visual grounding in gpt-4v.
\newblock \emph{arXiv preprint arXiv:2310.11441}.

\bibitem[{Yang et~al.(2019)Yang, Russakovsky, and Deng}]{yang2019spatialsense}
Kaiyu Yang, Olga Russakovsky, and Jia Deng. 2019.
\newblock Spatialsense: An adversarially crowdsourced benchmark for spatial relation recognition.
\newblock In \emph{Proceedings of the IEEE/CVF International Conference on Computer Vision}, pages 2051--2060.

\bibitem[{Yang et~al.(2020)Yang, Li, and Yu}]{yang2020graph}
Sibei Yang, Guanbin Li, and Yizhou Yu. 2020.
\newblock Graph-structured referring expression reasoning in the wild.
\newblock In \emph{Proceedings of the IEEE/CVF conference on computer vision and pattern recognition}, pages 9952--9961.

\bibitem[{Yarom et~al.(2023)Yarom, Bitton, Changpinyo, Aharoni, Herzig, Lang, Ofek, and Szpektor}]{yarom_what_2023}
Michal Yarom, Yonatan Bitton, Soravit Changpinyo, Roee Aharoni, Jonathan Herzig, Oran Lang, Eran Ofek, and Idan Szpektor. 2023.
\newblock \href {https://proceedings.neurips.cc/paper_files/paper/2023/file/056e8e9c8ca9929cb6cf198952bf1dbb-Paper-Conference.pdf} {What {You} {See} is {What} {You} {Read}? {Improving} {Text}-{Image} {Alignment} {Evaluation}}.
\newblock In \emph{Advances in {Neural} {Information} {Processing} {Systems}}, volume~36, pages 1601--1619. Curran Associates, Inc.

\bibitem[{Yildirim et~al.(2023)Yildirim, Baday, Erdem, Erdem, and Dundar}]{yildirim2023inst}
Ahmet~Burak Yildirim, Vedat Baday, Erkut Erdem, Aykut Erdem, and Aysegul Dundar. 2023.
\newblock Inst-inpaint: Instructing to remove objects with diffusion models.
\newblock \emph{arXiv preprint arXiv:2304.03246}.

\bibitem[{You et~al.(2024)You, Zhang, Gan, Du, Zhang, Wang, Cao, Chang, and Yang}]{you2024ferret}
Haoxuan You, Haotian Zhang, Zhe Gan, Xianzhi Du, Bowen Zhang, Zirui Wang, Liangliang Cao, Shih-Fu Chang, and Yinfei Yang. 2024.
\newblock \href {https://openreview.net/forum?id=2msbbX3ydD} {Ferret: Refer and ground anything anywhere at any granularity}.
\newblock In \emph{The Twelfth International Conference on Learning Representations}.

\bibitem[{Yu et~al.(2016)Yu, Poirson, Yang, Berg, and Berg}]{yu2016modeling}
Licheng Yu, Patrick Poirson, Shan Yang, Alexander~C Berg, and Tamara~L Berg. 2016.
\newblock Modeling context in referring expressions.
\newblock In \emph{Computer Vision--ECCV 2016: 14th European Conference, Amsterdam, The Netherlands, October 11-14, 2016, Proceedings, Part II 14}, pages 69--85. Springer.

\bibitem[{Yuksekgonul et~al.(2022)Yuksekgonul, Bianchi, Kalluri, Jurafsky, and Zou}]{yuksekgonul2022and}
Mert Yuksekgonul, Federico Bianchi, Pratyusha Kalluri, Dan Jurafsky, and James Zou. 2022.
\newblock When and why vision-language models behave like bags-of-words, and what to do about it?
\newblock In \emph{The Eleventh International Conference on Learning Representations}.

\bibitem[{Zhai et~al.(2023)Zhai, Yang, Zhao, Xu, Shen, Zhao, Keutzer, Li, Yan, and Fan}]{zhai2023halle-switch}
Bohan Zhai, Shijia Yang, Xiangchen Zhao, Chenfeng Xu, Sheng Shen, Dongdi Zhao, Kurt Keutzer, Manling Li, Tan Yan, and Xiangjun Fan. 2023.
\newblock Halle-switch: Rethinking and controlling object existence hallucinations in large vision language models for detailed caption.
\newblock \emph{arXiv preprint arXiv:2310.01779}.

\bibitem[{Zhang et~al.(2023)Zhang, Li, Li, Ren, Zou, Liu, Huang, Gao, Zhang, Li et~al.}]{zhang2023llava-g}
Hao Zhang, Hongyang Li, Feng Li, Tianhe Ren, Xueyan Zou, Shilong Liu, Shijia Huang, Jianfeng Gao, Lei Zhang, Chunyuan Li, et~al. 2023.
\newblock Llava-grounding: Grounded visual chat with large multimodal models.
\newblock \emph{arXiv preprint arXiv:2312.02949}.

\bibitem[{Zhang et~al.(2024)Zhang, Cai, Xie, and Lee}]{zhang-etal-2024-countercurate}
Jianrui Zhang, Mu~Cai, Tengyang Xie, and Yong~Jae Lee. 2024.
\newblock \href {https://doi.org/10.18653/v1/2024.findings-acl.915} {{C}ounter{C}urate: Enhancing physical and semantic visio-linguistic compositional reasoning via counterfactual examples}.
\newblock In \emph{Findings of the Association for Computational Linguistics: ACL 2024}, pages 15481--15495, Bangkok, Thailand. Association for Computational Linguistics.

\bibitem[{Zhao et~al.(2024{\natexlab{a}})Zhao, Zhao, G, Suh, Metaxas, Chandraker, and Schulter}]{Zhao_2024_CVPR}
Shiyu Zhao, Long Zhao, Vijay Kumar~B G, Yumin Suh, Dimitris~N. Metaxas, Manmohan Chandraker, and Samuel Schulter. 2024{\natexlab{a}}.
\newblock Generating enhanced negatives for training language-based object detectors.
\newblock In \emph{Proceedings of the IEEE/CVF Conference on Computer Vision and Pattern Recognition (CVPR)}, pages 13592--13602.

\bibitem[{Zhao et~al.(2024{\natexlab{b}})Zhao, Chen, Xu, Li, Wang, Li, and Huang}]{zhao2024open}
Xiangyu Zhao, Yicheng Chen, Shilin Xu, Xiangtai Li, Xinjiang Wang, Yining Li, and Haian Huang. 2024{\natexlab{b}}.
\newblock An open and comprehensive pipeline for unified object grounding and detection.
\newblock \emph{arXiv preprint arXiv:2401.02361}.

\bibitem[{Zhuang et~al.(2023)Zhuang, Zeng, Liu, Yuan, and Chen}]{zhuang2023task}
Junhao Zhuang, Yanhong Zeng, Wenran Liu, Chun Yuan, and Kai Chen. 2023.
\newblock A task is worth one word: Learning with task prompts for high-quality versatile image inpainting.
\newblock \emph{arXiv preprint arXiv:2312.03594}.

\end{thebibliography}

\appendix

\section{Dataset details}
\label{sec:appendix_dataset}

\begin{table*}[t]
\centering
\resizebox{0.9\textwidth}{!}{
\begin{tabular}{lp{0.4\textwidth}p{0.4\textwidth}} 
\toprule
\textbf{Type}             & \textbf{Exemplar templates}                                                  & \textbf{Expression examples}                                                                     \\ 
\midrule
0\_hop           & The <$att_0$> <$obj_0$>.                                                      & The white plate.                                                                        \\
1\_hop           & The <$att_0$> <$obj_0$> is <$rel_0$> the <$att_1$> <$obj_1$>.                              & The giraffe is to the right of the trees.                                             \\
and              & The <$att_0$> <$obj_0$> <$rel_0$> the <$att_1$> <$obj_1$> and <$rel_1$> the <$att_2$> <$obj_2$>.        & The balding man wearing the green shirt and to the left of the green trees.           \\
2\_hop           & The <$att_0$> <$obj_0$> is <$rel_0$> the <$att_1$> <$obj_1$> that is <$rel_1$> <$att_2$> <$obj_2$>.     & The blue, colorful and running train is on the bridge that is behind the green tree.  \\
same\_attr       & The <$obj_0$> sharing the <$rel_0$> as the <$obj_1$>.                            & The plate that has the same color as the rice.                                          \\
same\_attr\_2hop & The <$obj_0$> that has the <$rel_0$> as the <$obj_1$> that <$rel_1$> the <$att_0$> <$obj_2$>. & The table sharing the same color as the towels that to the right of the robe.         \\
\bottomrule
\end{tabular}}
\caption{\label{expression_template}
Examples of expression type. $obj_0$ denotes the target object, while $obj_{1,2}$ denote the related objects. $att_{0,1,2}$ and $rel_{0,1}$ denote the corresponding attributes and relations, respectively.
}
\end{table*}

\subsection{Predefined templates}
\label{sec:appendix_dataset_0}
We have meticulously crafted a variety of templates tailored to suit different sentence structures, encompassing a range of 1-3 templates per structure. Examples of templates and corresponding expressions are shown in Table~\ref{expression_template}.

\begin{table}
\centering
\begin{tabular}{lcccc} 
\toprule
\textbf{Set}   & \textbf{L1} & \textbf{L2} & \textbf{L3} & \textbf{Sum.}  \\ 
\midrule
Train & 134466      & 25282       & 4044        & 163792         \\
Test  & 5730        & 3404        & 471         & 9605           \\
Val   & 15126       & 2884        & 445         & 18455          \\
\bottomrule
\end{tabular}
\caption{\label{benchmark_cnt_pos}
Positive expressions Statistics. FineCops-Ref covers 3 difficult levels of positive expressions, split into train/test/val.
}
\end{table}

\begin{table}
\centering
\resizebox{1\columnwidth}{!}{
\begin{tabular}{lcccccc} 
\toprule
               & \multicolumn{3}{c}{\textbf{REPLACE}}                     & \multicolumn{2}{c}{\textbf{SWAP}}    &                \\ 
\cmidrule(lr){2-4}
\cmidrule(lr){5-6}
\textbf{Set}   & \textbf{Object} & \textbf{Attribute} & \textbf{Relation} & \textbf{Object} & \textbf{Attribute} & \textbf{Sum.}  \\ 
\midrule
Train & 29287           & 20678              & 14825             & 10062           & 5599               & 80451          \\
Test  & 3951            & 1725               & 1891              & 1722            & 525                & 9814           \\
Val   & 3308            & 2344               & 1676              & 1070            & 631                & 9029           \\
\bottomrule
\end{tabular}}
\caption{\label{benchmark_cnt_negtext}
Hard negative expressions Statistics. FineCops-Ref covers 5 fine-grained types of hard negative expressions, split into train/test/val.
}
\end{table}

\begin{table}
\centering
\resizebox{1\columnwidth}{!}{
\begin{tabular}{lcccccc} 
\toprule
             & \multicolumn{2}{c}{\textbf{REPLACE}} & \multicolumn{3}{c}{\textbf{SWAP}}                                      &                \\ 
\cmidrule(lr){2-3}\cmidrule(lr){4-6}
\textbf{Set} & \textbf{Object} & \textbf{Attribute} & \textbf{\textbf{Object}} & \textbf{\textbf{Attribute}} & \textbf{Flip} & \textbf{Sum.}  \\ 
\midrule
Test         & 4171            & 1844               & 307                      & 630                         & 1555          & 8507           \\
\bottomrule
\end{tabular}}
\caption{\label{benchmark_cnt_negimg}
Hard negative images Statistics. FineCops-Ref covers 5 fine-grained types of hard negative images.
}
\end{table}

\begin{figure*}
  \centering
  \includegraphics[width=\textwidth]{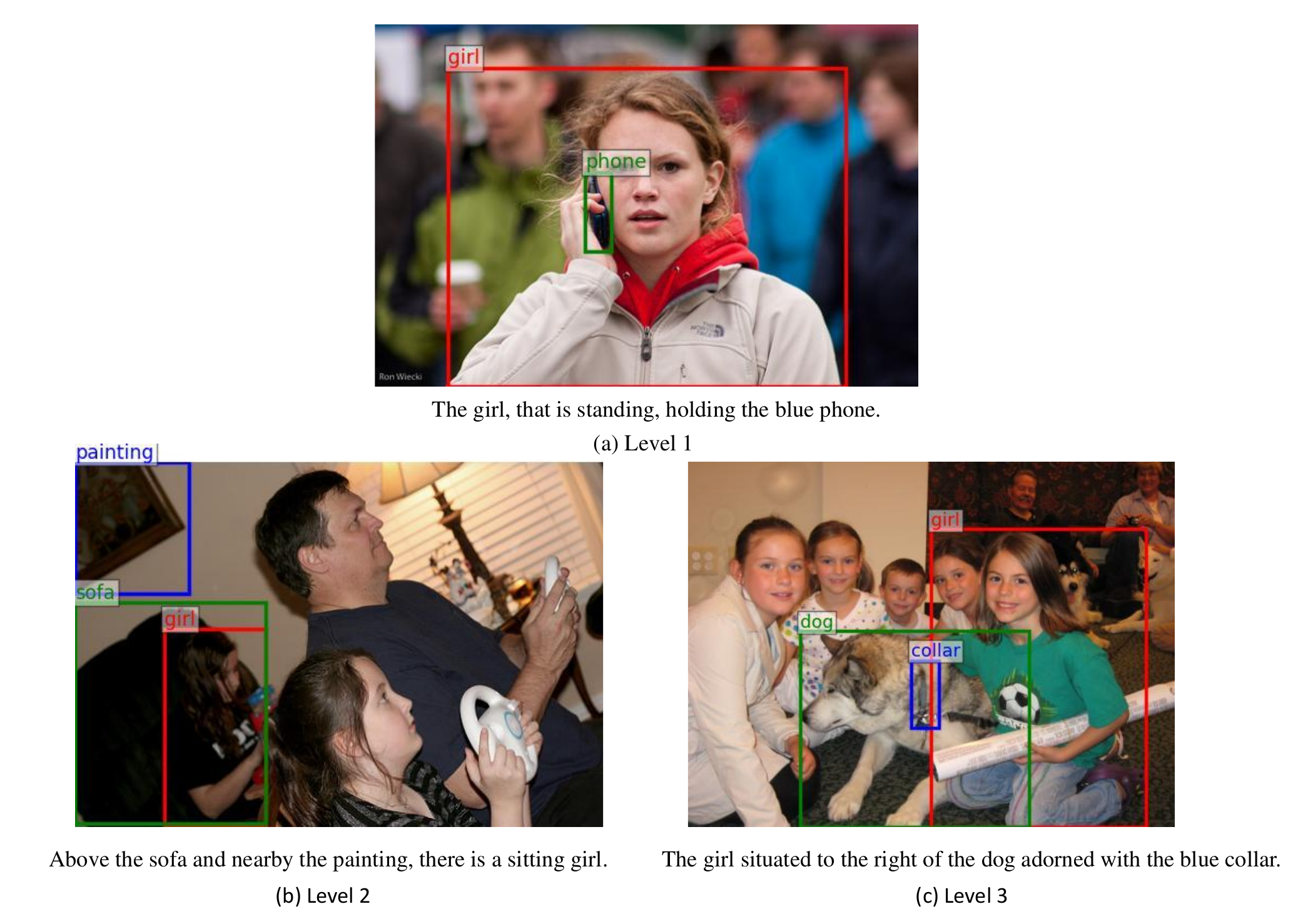}
  \caption{Positive expressions of different difficulty levels.}
  \label{level}
\end{figure*}

\begin{figure*}
  \centering
  \includegraphics[width=\textwidth]{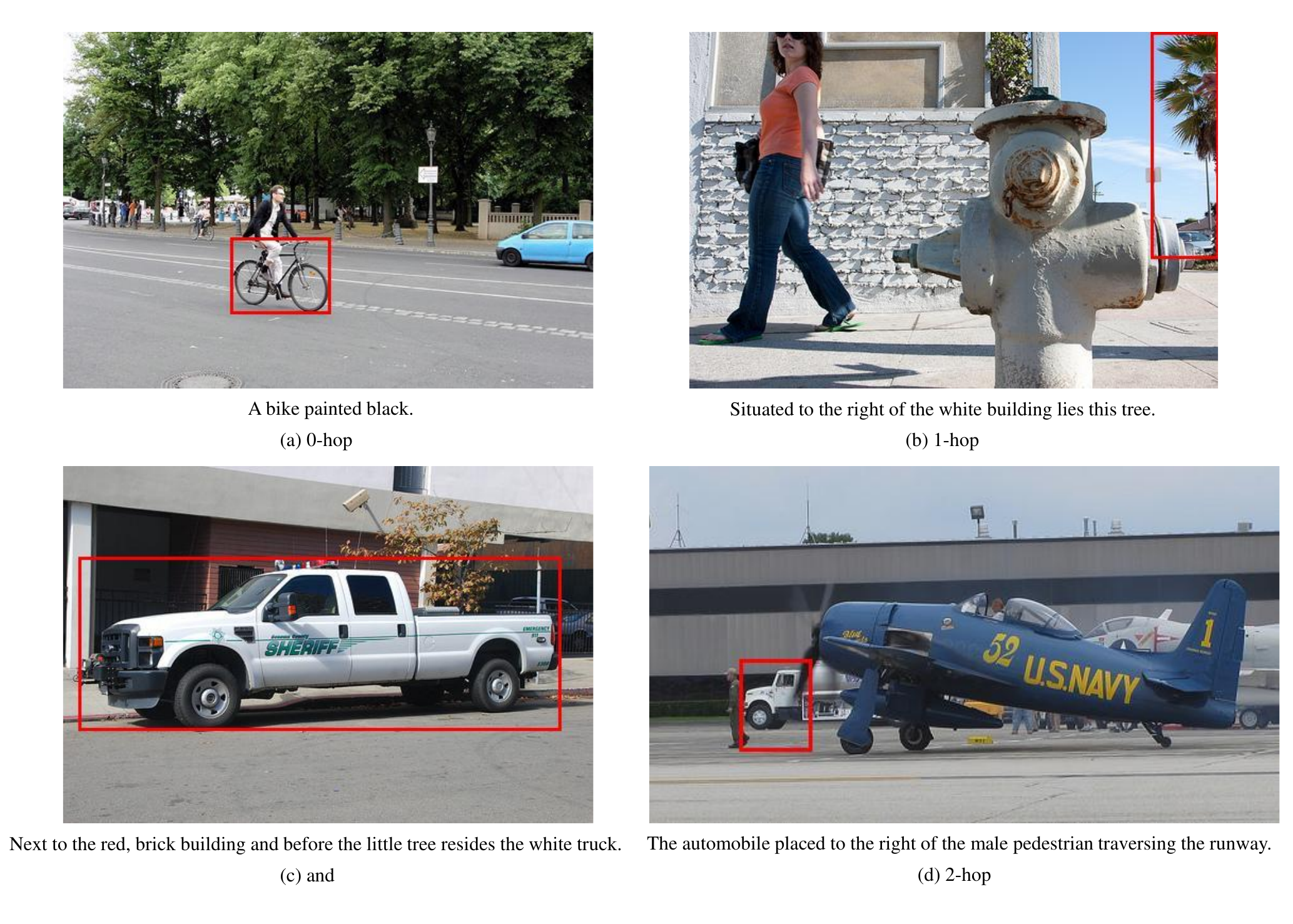}
  \caption{Positive expressions of different syntactic structure types.}
  \label{type}
\end{figure*}

\begin{figure*}
  \centering
  \includegraphics[width=\textwidth]{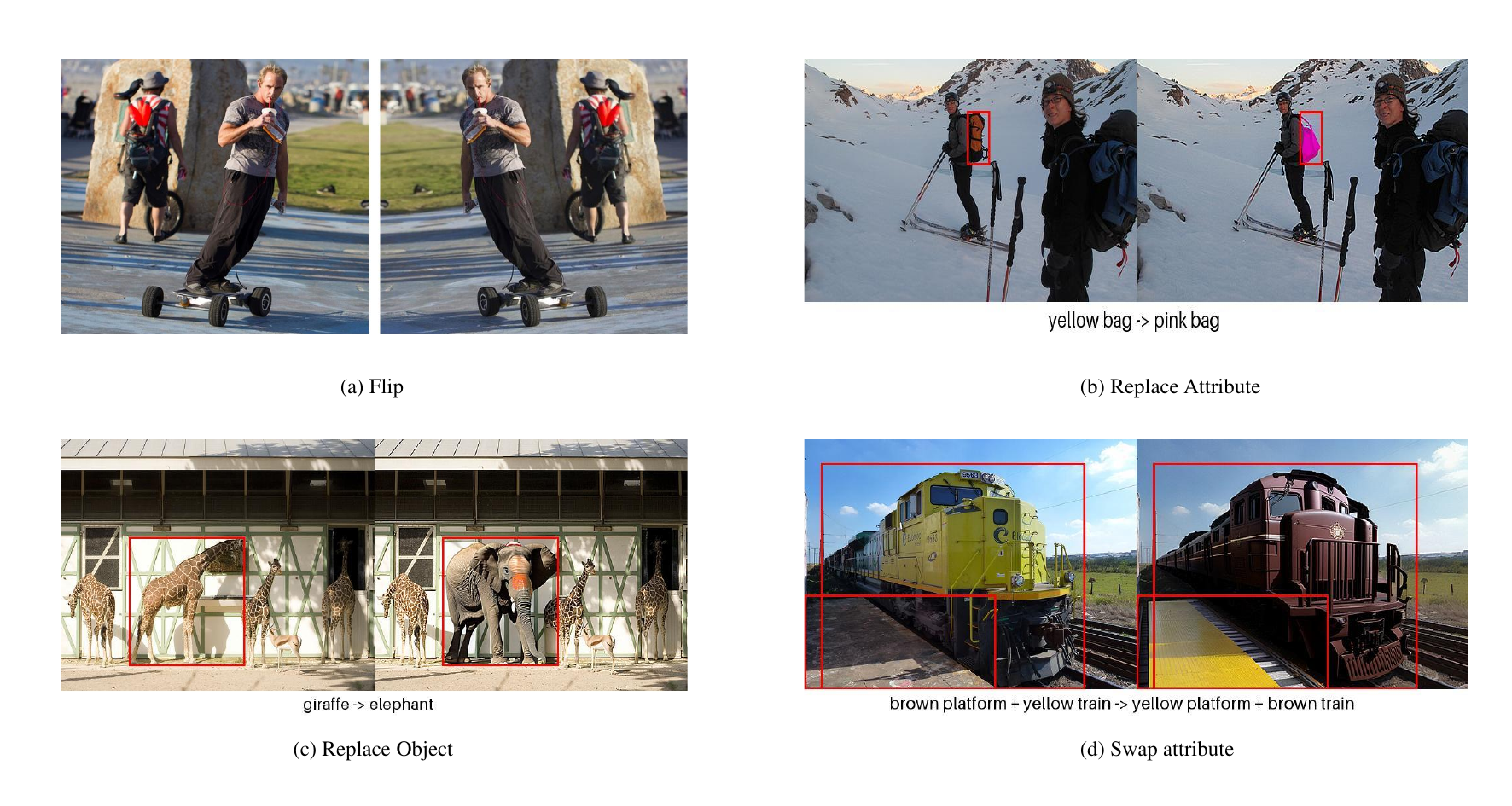}
  \caption{Negative images generated by different methods.}
  \label{neg_img}
\end{figure*}

\subsection{Examples of dataset}
\label{sec:appendix_dataset_1}

\textbf{Difficulty levels.}
We categorize positive expressions into three levels, depending on the complexity of fine-grained reasoning. The difficulty criterion is established based on the intricacy of fine-grained reasoning, rather than the complexity of the textual description. Figure~\ref{level} showcases exemplary data ranging in difficulty levels.

\textbf{Syntactic structure types.}
Meanwhile, following the syntactic structure, we categorize regular expressions into six types. $obj_0$ represents the target object, while $obj_1$ and $obj_2$ represent the related objects. "0\_hop" indicates that the expression only involves $obj_0$, "1\_hop" indicates that the expression mentions both $obj_0$ and $obj_1$. "And" and "2\_hop" encompass $obj_{0,1,2}$. In "and," $obj_1$ and $obj_2$ are in a coordinated relationship, whereas in "2\_hop," they are in a progressive relationship. "Same\_attr" and "same\_attr\_2hop" restrict the relationship between $obj_0$ and $obj_1$ to the same attribute. Figure~\ref{type} showcases exemplary data ranging in syntactic structure types.

\textbf{Negative images.}
Figure~\ref{neg_img} illustrates negative images generated by different methods.
\textbf{Dataset statistics.}
For positive expressions and negative expressions, we split the dataset into train, test, and val sets. Specifically, positive expressions are classified based on levels, as detailed in Table~\ref{benchmark_cnt_pos}. Negative expressions are classified based on types, as detailed in Table~\ref{benchmark_cnt_negtext}. While for negative images, we only generated them in the test set, categorized by type. Refer to Table~\ref{benchmark_cnt_negimg} for more details.

\subsection{Method to generate negative expressions}
\label{sec:appendix_dataset_2}
During our exploration into generating negative expressions, we delved into various methods to enhance the process. These methods encompassed the following approaches:
(1) Predefined replace list: This method involves utilizing a predefined list of replacement words to substitute specific words. Although simple, it suffers from limited diversity and substantial bias.
(2) Bert fill-mask: Employing this technique involves masking the original word and employing Bert to fill in the replacement. However, this method proves to be unstable and does not guarantee that the original word and its replacement belong to the same category.
(3) LLM replace: This approach prompts the Language Model to generate the replacement word. It offers a high degree of richness and delivers reasonable outputs. Nonetheless, it requires a significant amount of time.
In Table~\ref{generate_neg_exp}, we compare the outputs of these three methods using the vera and grammar score. The results indicate that LLM replace emerges as the optimal choice, both grammatically and logically.

\begin{table}
\centering
\begin{tabular}{lcc} 
\toprule
\textbf{Method}         & \textbf{Vera} & \textbf{Grammar}  \\ 
\midrule
Predefined replace list & 70            & 55                \\
Bert fill-mask          & 57            & 40                \\
LLM replace             & 61            & 50                \\
\bottomrule
\end{tabular}
\caption{\label{generate_neg_exp}
Vera and Grammar score of different method's output. The closer the score is to 50, the higher the quality of the data.
}
\end{table}

\subsection{Examples of prompt}
\label{sec:appendix_dataset_3}
\begin{figure}
  \centering
  \includegraphics[width=\columnwidth]{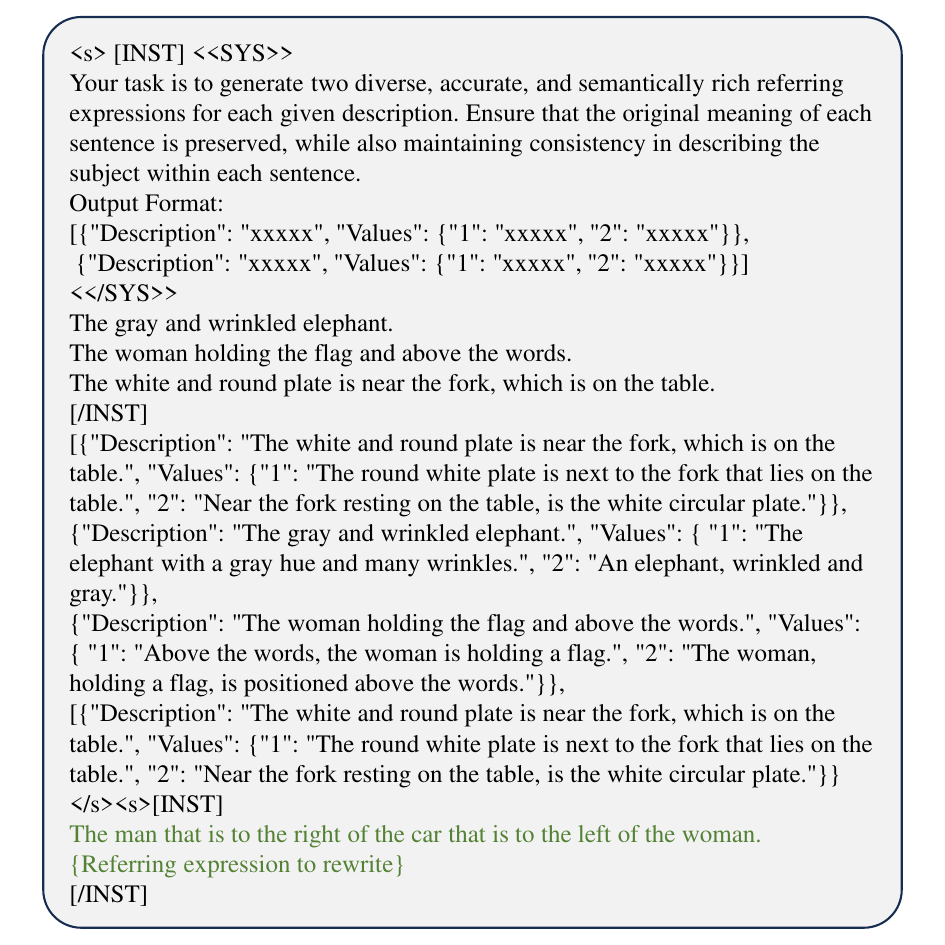}
  \caption{Prompt used for rewriting expressions.}
  \label{rewrite}
\end{figure}

\textbf{Prompt to rewrite expressions.}
We encourage the LLM to rephrase the given statement, aiming for rich and organic expressions while ensuring consistency throughout. Context learning was employed to integrate manually rewritten examples into the prompt. Additionally, to address any potential hallucinations, the LLM was instructed to include the original expression once in the output. Furthermore, two additional expressions were generated to enhance the diversity of the results. Figure~\ref{rewrite} illustrates the prompt provided to the LLM for the rewriting task.

\begin{figure*}
  \centering
  \includegraphics[width=\textwidth]{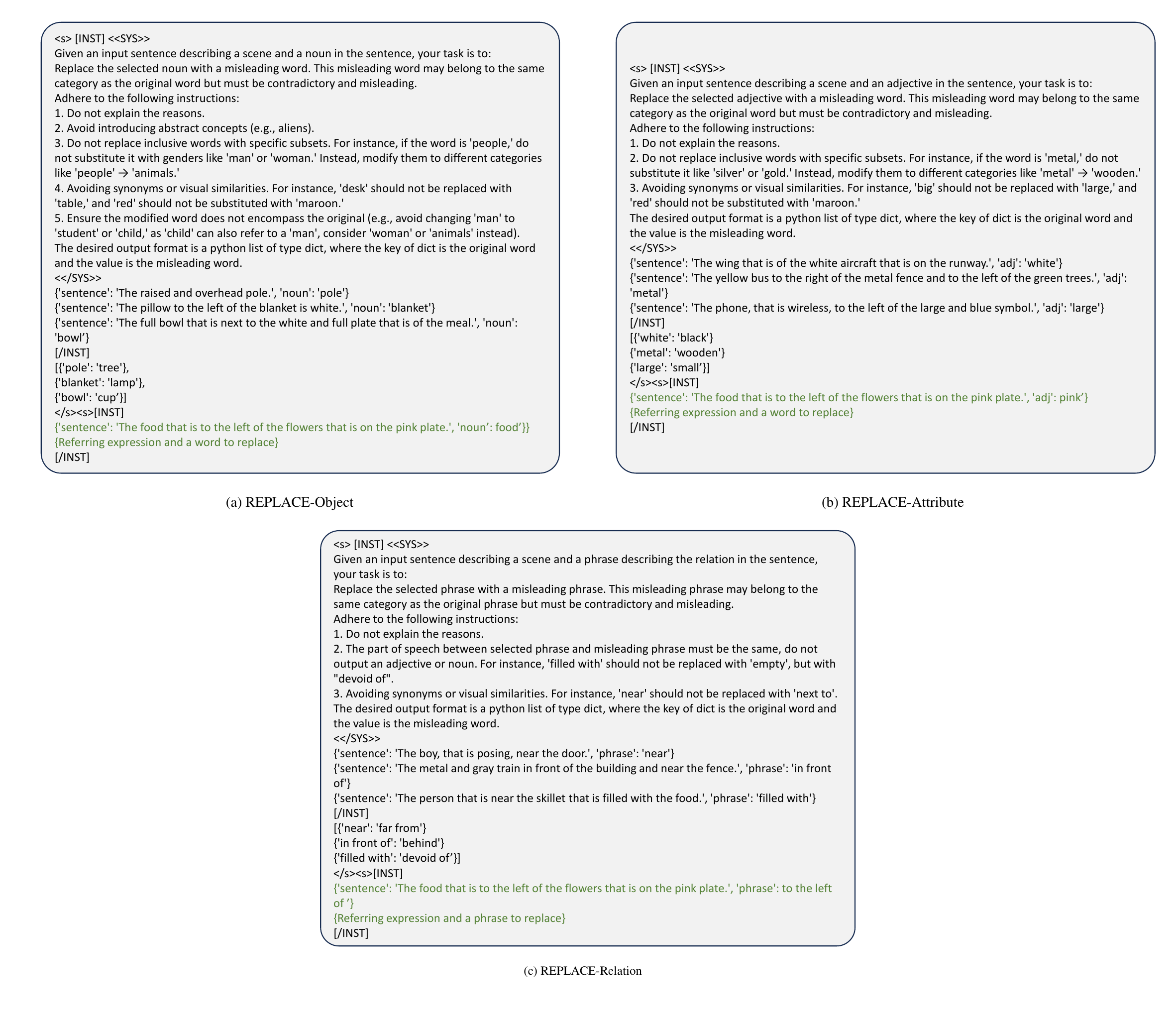}
  \caption{Prompt used for generating REPLACE negative expressions.}
  \label{replace}
\end{figure*}

\textbf{Prompt to generate negative expressions.}
We prompt LLM to replace the specified word in the expression. LLM is required to find a misleading word that falls within the same category as the original word, yet contradicts its meaning. To ensure consistent and precise results, we have implemented stringent guidelines within the prompt. Furthermore, we have employed context-based learning by incorporating manually replaced instances in the prompt. Figure~\ref{replace} illustrates the prompt provided to LLM for finding misleading words.

\begin{figure*}
  \centering
  \includegraphics[width=\textwidth]{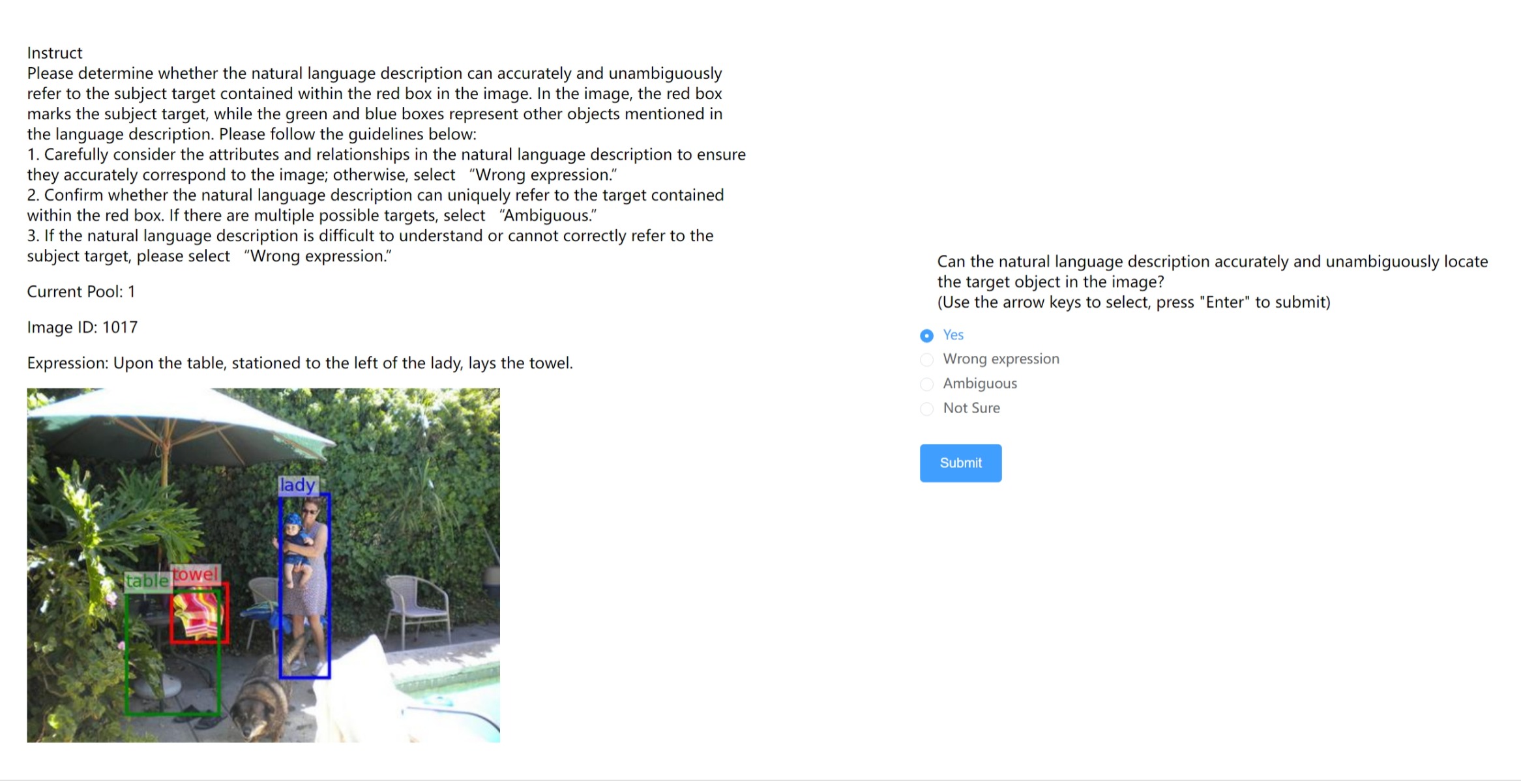}
  \caption{Program used for human filter.}
  \label{label_html}
\end{figure*}

\subsection{Human filter}
\label{sec:appendix_dataset_4}
We use the following prompt to guide human annotators to filter data. Program used for human filter see Figure~\ref{label_html}.

Please determine whether the natural language description can accurately and unambiguously 
refer to the subject target contained within the red box in the image. In the image, the red box 
marks the subject target, while the green and blue boxes represent other objects mentioned in 
the language description. Please follow the guidelines below: 

1. Carefully consider the attributes and relationships in the natural language description to ensure 
they accurately correspond to the image; otherwise, select “Wrong expression,” 

2. Confirm whether the natural language description can uniquely refer to the target contained 
within the red box. If there are multiple possible targets, select “Ambiguous,” 

3. If the natural language description is difficult to understand or cannot correctly refer to the 
subject target, please select "Wrong expression.”

\section{Implementation details}
\label{sec:appendix_exp}

\subsection{Hardware information}
All experiments are run on a machine with an Intel(R) Xeon(R) Gold 6348 CPU with a 512G
memory and four 80G NVIDIA RTX A800 GPUs.

\subsection{Dataset sources}
We obtain all existing datasets from their original sources released by the authors. We refer readers to these sources for the dataset licenses. To the best of our knowledge, the data we use does not contain personally identifiable information or offensive content.

\begin{itemize}
  \item GQA~\cite{hudson2019gqa}: We obtain GQA dataset from its official repository~\footnote{\url{https://cs.stanford.edu/people/dorarad/gqa/}}.
  \item RefCOCO~\cite{yu2016modeling}: We obtain RefCOCO dataset from its official repository~\footnote{\url{https://cocodataset.org/}}.
\end{itemize}

\subsection{Model configuration}

\textbf{Model sources.}
We detail the sources of the pretrained models we use in the paper.
\begin{itemize}
  \item MDETR~\cite{kamath2021mdetr}: We obtain MDETR from its official repository~\footnote{\url{https://github.com/ashkamath/mdetr}}. We use the refcocog\_EB3\_checkpoint.
  \item MM-GDINO~\cite{liu2023grounding}: We obtain MM-GDINO from its official repository~\footnote{\url{https://github.com/open-mmlab/mmdetection}}.
  \item UNINEXT-H~\cite{yan2023universal}: We obtain UNINEXT from its official repository~\footnote{\url{https://github.com/MasterBin-IIAU/UNINEXT}}.
  \item Shikra-7B~\cite{chen2023shikra}: We obtain Shikra from its official repository~\footnote{\url{https://github.com/shikras/shikra}}.
  \item Ferret-13B~\cite{you2024ferret}: We obtain Ferret from its official repository~\footnote{\url{https://github.com/apple/ml-ferret}}.
  \item GroundingGPT-7B~\cite{li-etal-2024-groundinggpt}: We obtain GroundingGPT from its official repository~\footnote{\url{https://github.com/lzw-lzw/GroundingGPT}}.
  \item Lenna-7B~\cite{wei2023lenna}: We obtain Lenna from its official repository~\footnote{\url{https://github.com/Meituan-AutoML/Lenna}}.
  \item CogVLM-grounding-generalist-17b~\cite{wang2023cogvlm}: We obtain CogVLM from its official repository~\footnote{\url{https://github.com/THUDM/CogVLM}}.
  \item CogCoM-grounding-17b~\cite{qi2024cogcom}: We obtain CogCom from its official repository~\footnote{\url{https://github.com/THUDM/CogCoM}}.
  \item GPT-4 Turbo~\footnote{\url{https://platform.openai.com/docs/models}}: We ues GPT-4 via API. The version is gpt-4-turbo-2024-04-09.
\end{itemize}

\subsection{Experiments details}

\textbf{Evaluation details.}
We obtain the bounding box coordinates and confidence scores predicted by the model on our benchmark, and then calculate the metrics.
\begin{itemize}
    \item \textbf{Specialist:} We use the official inference code to perform inference and record the bounding box coordinates and confidence scores of the output.
    \item \textbf{MLLMs:} We use the official inference code for inference and record the bounding box coordinates. The confidence score is calculated using the sum of the log probabilities of the coordinate tokens~\cite{kurita2023refego,mitchell2023detectgpt}.
    \item \textbf{GPT-4V+SoM:} Following the SoM~\cite{yang2023set}, we first use MM-GDINO to obtain candidate bounding boxes. Then, we draw these bounding boxes and corresponding labels on the image and ask GPT-4v to choose the label. To save costs, testing was conducted on a sample of 5k instances.
\end{itemize}

\begin{table*}[t]
\centering
\resizebox{0.8\textwidth}{!}{
\begin{tabular}{lccccccccccc} 
\toprule
                    & \multicolumn{6}{c}{\textbf{REPLACE}}                                                                                 & \multicolumn{4}{c}{\textbf{SWAP}}                                            &                 \\ 
\cmidrule(r){2-7}\cmidrule(lr){8-11}
                    & \multicolumn{2}{c}{\textbf{Object}} & \multicolumn{2}{c}{\textbf{Attribute}} & \multicolumn{2}{c}{\textbf{Relation}} & \multicolumn{2}{c}{\textbf{Object}} & \multicolumn{2}{c}{\textbf{Attribute}} &                 \\ 
\cmidrule(lr){2-3}\cmidrule(lr){4-5}\cmidrule(lr){6-6}\cmidrule(lr){7-7}\cmidrule(lr){8-9}\cmidrule(lr){10-11}
\textbf{Model}      & L1             & L2                 & L1             & L2                    & L1             & L2                   & L1             & L2                 & L1             & L2                    & \textbf{Avg.}   \\ 
\midrule
\textbf{Specialist} &                &                    &                &                       &                &                      &                &                    &                &                       &                 \\
MDETR               & 63.58          & 51.89              & 58.75          & 52.64                 & 54.92          & \uline{54.26}        & 59.60          & 54.11              & 56.33          & 51.38                 & 55.75           \\
MM-GDINO-T          & 66.02          & 49.66              & 57.50          & 48.85                 & 49.88          & 49.78                & 56.80          & 49.50              & 55.87          & \uline{55.93}         & 53.98           \\
MM-GDINO-L          & 66.73          & 49.93              & 58.35          & 50.21                 & 51.93          & 53.57                & 60.51          & \uline{55.47}      & 54.88          & 54.72                 & 55.63           \\
UNINEXT             & 61.24          & 51.39              & 57.59          & 51.62                 & 54.35          & 52.22                & 58.57          & 52.05              & \uline{57.07}  & 49.58                 & 54.57           \\
MM-GDINO-T$\dagger$      & 71.00          & 51.80              & 57.68          & 49.33                 & 53.23          & 50.42                & \uline{63.57}  & 53.75              & 56.89          & 49.26                 & 55.69           \\
MM-GDINO-T$\ddagger$      & \textbf{80.84} & \textbf{70.86}     & \textbf{73.43} & \textbf{65.31}        & \textbf{70.85} & \textbf{67.22}       & \textbf{72.36} & \textbf{65.93}     & \textbf{71.70} & \textbf{75.75}        & \textbf{71.43}  \\ 
\midrule
\textbf{MLLM}       &                &                    &                &                       &                &                      &                &                    &                &                       &                 \\
Shikra              & 58.57          & 51.14              & 55.37          & 52.96                 & 52.67          & 52.88                & 57.07          & 51.44              & 55.04          & 48.42                 & 53.56           \\
Ferret-13B          & 52.44          & 49.34              & 49.39          & 48.80                 & 50.17          & 48.24                & 51.07          & 48.80              & 49.93          & 50.04                 & 49.82           \\
GroundingGPT        & 55.14          & 50.90              & 50.76          & 49.45                 & 50.04          & 48.15                & 53.11          & 49.44              & 49.83          & 50.51                 & 50.73           \\
Lenna               & \uline{76.46}  & \uline{63.93}      & \uline{64.29}  & 52.66                 & \uline{56.92}  & 53.56                & 59.98          & 51.22              & 56.96          & 48.87                 & \uline{58.49}   \\
CogVLM              & 60.60          & 51.40              & 55.66          & \uline{52.96}         & 51.95          & 53.77                & 55.14          & 55.04              & 53.09          & 55.47                 & 54.51           \\
CogCom              & 63.47          & 52.51              & 56.83          & 52.60                 & 53.28          & 51.83                & 58.60          & 54.08              & 54.87          & 49.79                 & 54.79           \\
CogVLM$\dagger$        & 62.79          & 50.7               & 54.52          & 51.53                 & 51.72          & 51.16                & 55.22          & 53.97              & 50.79          & 50.55                 & 53.30           \\
\bottomrule
\end{tabular}}
\caption{\label{benchmark_text_auroc}
Evaluation results (AUROC) on negative expressions.
}
\end{table*}

\begin{table*}
\centering
\resizebox{0.8\textwidth}{!}{
\begin{tabular}{lcccccccccc} 
\toprule
                    & \multicolumn{4}{c}{\textbf{REPLACE}}                                         & \multicolumn{5}{c}{\textbf{SWAP}}                                                            &                 \\ 
\cmidrule(lr){2-5}\cmidrule(lr){6-10}
                    & \multicolumn{2}{c}{\textbf{Object}} & \multicolumn{2}{c}{\textbf{Attribute}} & \textbf{Object} & \multicolumn{2}{c}{\textbf{Attribute}} & \multicolumn{2}{c}{\textbf{Flip}} &                 \\ 
\cmidrule(lr){2-3}\cmidrule(lr){4-5}\cmidrule(lr){6-6}\cmidrule(lr){7-8}\cmidrule(r){9-10}
\textbf{Model}      & L1             & L2                 & L1             & L2                    & L1              & L1             & L2                    & L1             & L2               & \textbf{Avg.}   \\ 
\midrule
\textbf{Specialist} &                &                    &                &                       &                 &                &                       &                &                  &                 \\
MDETR               & 64.00          & 56.20              & 58.02          & 53.69                 & \uline{60.89}   & 58.72          & 55.63                 & \textbf{55.01} & \uline{53.42}    & 57.29           \\
MM-GDINO-T          & 64.32          & 57.51              & 53.27          & 55.81                 & 58.74           & 58.76          & 55.09                 & 51.72          & \textbf{53.43}   & 56.52           \\
MM-GDINO-L          & 68.00          & 58.05              & 55.90          & 56.08                 & 59.96           & \uline{62.81}  & 58.08                 & 51.87          & 53.04            & 58.20           \\
UNINEXT             & 62.13          & 53.92              & 54.80          & 52.44                 & \textbf{63.20}  & 57.49          & 50.76                 & 51.14          & 49.57            & 55.05           \\
MM-GDINO-T$\dagger$      & 70.03          & 58.22              & 57.71          & 55.71                 & 59.79           & 60.78          & 54.27                 & 51.25          & 51.48            & 57.69           \\
MM-GDINO-T$\ddagger$      & \textbf{75.07} & \uline{63.05}      & \textbf{65.20} & \textbf{61.35}        & 57.48           & \textbf{63.93} & \textbf{64.96}        & 51.65          & 51.59            & \textbf{61.59}  \\ 
\midrule
\textbf{MLLM}       &                &                    &                &                       &                 &                &                       &                &                  &                 \\
Shikra              & 55.94          & 50.40              & 50.92          & 51.36                 & 52.47           & 56.64          & 47.08                 & 51.57          & 51.45            & 51.98           \\
Ferret-13B          & 56.09          & 52.61              & 51.01          & 51.20                 & 55.78           & 53.80          & 49.49                 & 51.24          & 50.99            & 52.47           \\
GroundingGPT        & 56.75          & 50.86              & 48.52          & 49.37                 & 54.09           & 51.76          & 50.67                 & 52.84          & 47.46            & 51.37           \\
Lenna               & \uline{74.71}  & \textbf{65.17}     & \uline{60.27}  & 55.85                 & 59.24           & 59.08          & 52.47                 & 50.25          & 49.42            & \uline{58.50}   \\
CogVLM              & 58.62          & 55.88              & 51.03          & 55.24                 & 56.71           & 56.29          & 55.81                 & 52.04          & 51.47            & 54.79           \\
CogCom              & 37.91          & 33.34              & 31.45          & 29.67                 & 47.38           & 34.57          & 31.32                 & 33.44          & 31.56            & 34.52           \\
CogVLM$\dagger$        & 63.24          & 56.15              & 50.5           & \uline{56.86}         & 58.96           & 57.25          & \uline{59.49}         & \uline{53.09}  & 51.54            & 56.34           \\
\bottomrule
\end{tabular}}
\caption{\label{benchmark_img_auroc}
Evaluation results (AUROC) on negative images.
}
\end{table*}

\textbf{Training detials.}
We detail the dataset and hyper-parameters used in training our own models.

\begin{itemize}
    \item \textbf{MM-GDINO-T:} We trained the model with a batch size of 32. The AdamW optimizer was used with a learning rate of 0.0002 and a weight decay of 0.0001. The learning rate was adjusted using a MultiStepLR scheduler. The training ran for 5 epochs. For negative samples, the ground truth bounding box was set as empty.
    \item \textbf{CogVLM:} We followed the provided template and performed instruction tuning with the joined training set of ours and RefCOCO/+/g. The training was done with lora~\cite{hu2022lora} and a batch size of 32, using the AdamW optimizer with a learning rate of 0.0002 and a weight decay of 0.0001. The training ran for 1 epoch, with a cosine learning rate schedule.
\end{itemize}

\section{Detailed evaluation results}
\label{sec:appendix_results}

\textbf{AUROC results.}
The experimental results of AUROC are shown in Table~\ref{benchmark_text_auroc} and \ref{benchmark_img_auroc}, which exhibit a similar trend to Recall, further confirming the following observations: (1) The models are highly sensitive to the specific locations of negative data. (2) The models have a poor understanding of relationships. 
\end{document}